\documentclass[journal]{IEEEtran}
\ifCLASSINFOpdf
\else
\fi
\usepackage{subcaption}

\usepackage{graphicx}
\usepackage{amsmath}
\usepackage{multirow}
\usepackage{color,soul}

\begin{document}
%
\title{A clustering approach to time series forecasting using neural networks: A comparative study on distance-based vs. feature-based clustering methods.}
%
%
%

\author{Manie Tadayon
                \hspace{6mm} Yumi Iwashita
\thanks{Manie Tadayon is with ECE Department at UCLA and Robotic group at Jet Propulsion Laboratory, Pasadena, CA 91101 (e-mail: manie.tadayon@jpl.nasa.gov)}
\thanks{Yumi Iwashita is with Robotic group at Jet Propulsion Laboratory, Pasadena, CA 91101 (e-mail: yumi.iwashita@jpl.nasa.gov)}}
\maketitle

\begin{abstract}
Time series forecasting has gained lots of attention recently; this is because many real-world phenomena can be modeled as time series. The massive volume of data and recent advancements in the processing power of the computers enable researchers to develop more sophisticated machine learning algorithms such as neural networks to forecast the time series data. In this paper, we propose various neural network architectures to forecast the time series data using the dynamic measurements; moreover, we introduce various architectures on how to combine static and dynamic measurements for forecasting. We also investigate the importance of performing techniques such as anomaly detection and clustering on forecasting accuracy. Our results indicate that clustering can improve the overall prediction time as well as improve the forecasting performance of the neural network. Furthermore, we show that feature-based clustering can outperform the distance-based clustering in terms of speed and efficiency. Finally, our results indicate that adding more predictors to forecast the target variable will not necessarily improve the forecasting accuracy.
\end{abstract}


%
\IEEEpeerreviewmaketitle

\section{Introduction}
%
%
%
%
\IEEEPARstart{T}{ime} series forecasting has received lots of attention recently. This is because many phenomena such as stock price, temperature, and weather can be modeled as time series. The fundamental challenge in time series forecasting is that the observations at different points in time are correlated, which makes some of the algorithms that change or permute the order of observations unusable.

Scientists and researchers have done extensive research in time series forecasting, such as \cite{hassan2005stock,tadayon2019predicting,ariyo2014stock,contreras2003arima,ediger2007arima}. They have borrowed tools from various domains, such as graphical modeling and statistics, to improve forecasting accuracy. For example, in \cite{hassan2005stock} and \cite{tadayon2019predicting}, the authors used the hidden Markov model (HMM) to predict time series data. In \cite{tadayon2019predicting}, the authors showed that the HMM can achieve high accuracy to predict student performance in an educational game. In \cite{ariyo2014stock,contreras2003arima} and \cite{ediger2007arima}, the authors used the well-known time series method, namely, autoregressive integrated moving average (ARIMA) to make the forecast of the stock market, electricity prices and energy demands.

Although classical methods like ARIMA and HMM proved to be successful in modeling time-series data, however, they exhibit some limitations as follows: 1- HMM follows the one step Markov assumption, which means conditioning on the current state the future state is independent of all the past states. This assumption might be violated for most real-world time series data. 2- HMM considers there are underlying hidden states that are responsible to generate the observations, which might not be correct for many time series forecasting problems. 3- ARIMA and most other classical time series models can only be used for univariate time series; furthermore, it is shown that they cannot model nonlinear time series accurately. \cite{aladag2009forecasting,zhang2003time,chakraborty1992forecasting}. 

Recent advancements in the processing power of the computers enable researchers to take advantage of the massive volume of data and develop more sophisticated algorithms such as neural networks to model and predict the data. Many researchers have used neural networks in the time series forecasting and showed that they can lead to better prediction than ARIMA model. \cite{chakraborty1992forecasting,zhang2003time,siami2018comparison}. A few papers such as \cite{tadayon2020comparative} used a simulative approach to compare the HMM and the LSTM for various dynamic Bayesian network structures. They showed that the LSTM outperforms the HMM when the dataset is large.

Time series clustering is another major topic that receives lots of attention. Extensive research such as \cite{aghabozorgi2015time,liao2005clustering} and \cite{oates1999clustering} is done in developing accurate and efficient clustering algorithms. In \cite{aghabozorgi2015time} and \cite{liao2005clustering}, the authors performed the comprehensive surveys of the time series clustering methods. They described distance-based clustering methods such as dynamic time warping (DTW) as well as feature-based and representation based clustering methods.  
In \cite{bandara2020forecasting}, the authors combined the recurrent neural network, specifically long short term memory (LSTM) and feature-based clustering to improve the time series forecasting accuracy. In particular, they showed that their method outperforms the baseline LSTM model in terms of mean sMAPE accuracy. 

Some papers such as \cite{yen2019integration,leontjeva2016combining} and \cite{hsu2019enhanced} made forecasting using the neural network by combining static and dynamic features. They showed that combining static and dynamic features can outperform the classification accuracy using the static or dynamic features alone. 

In this paper, we are proposing to combine anomaly detection, clustering, and forecasting using LSTM for time series data to achieve better prediction. The contribution of this paper is fourfold. First, we perform anomaly detection to identify and replace the outliers in the time series data. This is a major preprocessing step that can significantly affect the prediction and clustering.
Second, we introduce some methods to impute the missing values for time series data. Third, we compare distance-based and feature-based clustering in terms of speed and prediction accuracy. Fourth, we introduce multiple architectures to make forecasting using time series data as well as multiple architectures to combine static data and time series data.

The rest of this article is organized as follows: Section II defines the methods and algorithms employed in the later sections. Section III describes the dataset used in the article. Section IV describes the problem formulation. Section V presents and discusses the results. Finally, section VI presents our conclusion.

\section{Algorithms and Methods}
\subsection{Anomaly Detection}
Anomaly detection is an essential step before applying any machine learning algorithm. This step should be taken before imputing the missing values since the outliers will influence the missing values. Outliers are defined as observations that are significantly different from all other observations in data. If they are not detected and replaced correctly, they will have a negative impact on clustering and forecasting outcomes.

Several algorithms, such as \cite{malhotra2015long} and \cite{pincombe2005anomaly} are proposed to detect anomalies in time series using the ARMA process and the neural network, specifically LSTM. In this paper, we are using seasonality and trend decomposition based on locally estimated scatterplot smoothing (Loess) to detect the outliers in time series. This is the algorithm developed by Twitter and works by decomposing the time series to seasonality, trend, and random components. It is suitable for seasonal time series and is claimed to work accurately for additive outliers. An example of the implementation of this algorithm can be found in the tsclean package at R. 
\subsection{Missing Value Imputation}
The second major preprocessing step after detecting and replacing the outliers is imputing the missing values. Depending on the nature of data, several methods exist to impute the missing values. Following techniques are examples of missing value imputation for time series data:
\begin{table}[h]
\centering
\caption{\label{tab:1} Time Series Missing Value Imputation Methods}
\begin{tabular}{ |l|l| }
  \hline
  \multicolumn{2}{|c|}{\textbf{Time Series Data}} \\
  \hline
  Interpolation & Linear, Spline, ... \\
  Locf & Last Observation Carry Forward \\
  Nocb & Next Observation Carry Backward \\
  Moving Average & Exponential,Linear, ... \\
  Mean & Mean, Median, ... \\
  \hline
\end{tabular}
\end{table}

Table 1 lists only some of the well-known methods to impute missing values. It is also worth mentioning that the tsclean package mentioned in the last section performs linear interpolation to impute missing values after anomaly detection. 
Several algorithms are proposed to impute missing values for the static data such as \cite{batista2002study,troyanskaya2001missing,mazumder2010spectral,hastie2015matrix}. In \cite{troyanskaya2001missing} the authors performed a comparative study of several missing value imputation techniques such as Singular Value Decomposition (SVD) based method (SVDimpute), weighted K-nearest neighbors (KNNimpute), and row average (filling missing values with zeros). They showed that KNNimpute provides a more robust and sensitive method than row average and SVDimpute. In \cite{mazumder2010spectral}, the authors proposed the SOFT-IMPUTE algorithm to iteratively replace the missing elements with those obtained from a soft-thresholded SVD. The softImpute Package in R described In \cite{hastie2015matrix} implements Iterative methods for matrix completion that use nuclear-norm regularization.

\subsection{Clustering}
Clustering is defined as dividing the data into some groups such that data in the same groups are more similar. It is divided into two major categories of distance-based and feature-based clustering. In distance-based clustering, the distance between data points is considered as similarity measure, and the objective is to have lower intra-cluster distance than inter-cluster distance. The main challenge in distance-based clustering is the notion of the distance itself. Choosing the appropriate distance is dependent on the nature of the dataset. In the case of static and cross-sectional data, euclidean distance is the most popular distance metric, and K-mean and hierarchical algorithms are two popular clustering choices. Hierarchical clustering is divided into two categories of agglomerative and divisive. Agglomerative hierarchical clustering is a more popular choice than divisive hierarchical clustering in which each data point is initially in its own cluster; then, at each iteration, the similar clusters merge with each other until K clusters are formed. The advantage of hierarchical clustering over K-mean is that we do not need to specify the number of clusters in advance. 

Dynamic time warping (DTW) is one of the most popular distance metrics used for time series clustering. It is a well-known technique that was originally developed for speech recognition application and is used to find an optimal alignment between two given time-dependent sequences. It is a dynamic programming algorithm that Unlike the Euclidean distance, is not susceptible to distortions in the time-axis. It works by warping the sequences in a nonlinear fashion to match each other \cite{muller2007dynamic}. Another advantage of DTW over euclidean distance is that it does not require the sequences to have the same lengths. However,despite its numerous applications, DTW suffers from quadratic computational complexity. This means if m and n represent the length of two sequences, then the computational complexity of finding the DTW distance between them is $O(m * n)$. Some papers such as \cite{silva2016speeding,salvador2007toward} proposed algorithms such as PrunedDTW, FastDTW to speedup the classical DTW algorithm.

Feature-based clustering works by first extracting features from data and then constructing the feature matrix from the extracted features. This method has several advantages over distance-based clustering. First, it is faster, as will be shown later in this paper. Second, standard clustering methods such as K-mean and hierarchical clustering can directly be applied to the feature matrix. 

In this paper, we are introducing two feature extraction methods for time series data. Method A mainly extracts time series specific features such as entropy, autocorrelation, partial autocorrelation, stability, and holt parameter.
These features showed to be very effective for clustering and detecting unusual time series \cite{wang2006characteristic,hyndman2015large}. 

Method B looks at the time series more like a signal and extracts a more exhaustive list of features such as energy, fast Fourier transform (FFT) coefficients, continuous wavelet transform (CWT) coefficients, variance \cite{christ2018time}.

\subsection{ Forecasting with Neural Networks}
Neural networks are drawing lots of attention in recent years. They have been used successfully in various domains such as signal and image processing, control, biology, and finance. Different neural network architectures are used depending on the application and the dataset. For example, if the objective is classification and the input data is a set of images, then a convolutional neural network (CNN) might be the best choice. If the input data is a time series, then recurrent neural network (RNN), specifically LSTM, is a better choice. However, sometimes more than one network type can be used in a given problem. For instance, a multilayer perceptron (MLP) and CNN are also used for time series forecasting. Time series should be reshaped in input and output format to be used by MLP. For example, assume a time series X with length n and the following elements:
$ X(1),X(2),X(3), ..., X(n-1),X(n)$. In order to use MLP to forecast the next value, X needs to be reshapes as follows:

\begin{table}[h]
\centering
\caption{\label{tab:2} Time Series Formulation For MLP}
\resizebox{3cm}{!}{
\begin{tabular}{ |c|c| } 
\hline
Input & Output \\
\hline
X(1) & X(2) \\ 
X(2) & X(3) \\
\vdots & \vdots \\
X(n-1) & X(n)  \\
\hline
\end{tabular}
}
\end{table}

RNN networks are mainly designed for sequence prediction problem. Unlike MLP, which considers input and output to be independent, RNN networks consist of memory cells that can remember the long term dependencies between elements of a sequence. In theory, RNN can remember arbitrary long time steps, but in practice, they suffer from vanishing gradient problem \cite{hochreiter1998vanishing,pascanu2013difficulty}. LSTM is designed to address the vanishing gradient problem. It consists of cell states and various gate elements that decide which data in a sequence is necessary to keep or throw away. By doing that, it can pass only relevant information down the long chain of sequences to make predictions.
\subsection{Error Metrics}
Forecasting error is divided into two categories of scale-dependent and scale-independent. Two well-known scale-dependent measures are mean absolute error (MAE), and root mean square error (RMSE).  Minimizing the MAE will lead to forecasts of the median while minimizing the RMSE will lead to forecasts of the mean. Although both MAE and RMSE are widely used error measures, MAE has the advantage that it is easier to compute and understand \cite{hyndman2018forecasting}. MAE and RMSE are defined as follows:
\begin{align}
    & e_{t}  = y_t - \hat{y}_t \\
     & MAE    = mean (|e_{t}|)  \\
    & RMSE  = \sqrt{mean(e_{t}^2)}
\end{align}
Mean absolute percentage error (MAPE) is a widely used scale-independent error measure that is defined as follows:
\begin{align}
   & P_{t}  = \dfrac{100 e_{t}}{y_{t}}\\
   &    MAPE =  mean (| P_{t} |) 
\end{align}
Despite being commonly used in practice, it has the following disadvantages:
1- It is undefined when the true value $y_{t}$ is zero, and it will be a very large value when $y_{t}$ is close to zero. 2- It puts a heavier weight on the negative errors than the positive ones. 3- It does not make sense when the measurement has an arbitrary zero point \cite{hyndman2018forecasting,hyndman2006another}.
\section{Dataset}
In this paper, We are using the synthetic data created as follows:
First, we employ ARIMA models with random AR and MA coefficients to generate multivariate time series with length L and a different range of values for different columns. Second, we find the absolute value of each column of the time series to make sure that each column consists of only positive values. Third, we add arbitrary number of columns to the previously generated columns in the last two steps by simulating the piecewise continuous function that takes constant value $C_1$ from 0 to $t_{1}$ and $C_2$ from $t_{1}$ to L. Fourth, replace $L_1$ elements of time series by big values larger than the maximum value of each column (measurement) and $L_2$ elements by zeros to simulate the effect of outliers. Finally, additive white Gaussian noise (AWGN) with a specific variance depending on the range of the data is added to each column. 

Above method has the following advantages: First, parameters of data such as the dimension of time series (number of measurements in multivariate time series), the variance of the noise, the minimum and the maximum value of time series corresponding to each measurement, and number of outliers can easily be defined and modified in simulation. Second, an arbitrary number of multivariate time series data can easily be created inside a loop.

Generating static data is more straightforward and is done as follows: First, identify the number of features as K. Second, determine the maximum and the minimum value and the type of each feature. Third, generate or sample data for each feature given the range of the values and the type of each feature.

For this paper, we are generating 400 multivariate time series data with three measurement columns. The total number of outliers per each measurement is 10. The length of each measurement is 400 as well. We generate 400 static data with five continuous features. Therefore the static data is a matrix of 400 by 5. 

For the multivariate time series, we will be using generic names. For example, the first measurement is oil, then is water, and the third column is gas. For static features, we will call them feature 1, feature 2, ... , feature 5.

All the simulations run on Mac pro laptop with 16 GB memory and 2.80 GHz CPU.
\section{Problem Formulation}
In this paper, The objective is to predict the cumulative gas value (Third column of time series) given all other data and measurements. We use the first N elements of time series and the static data for training and try to predict the $K^{th}$ cumulative gas value where $K > N$. More specifically, we use $N=100$ and K=150, 200, 300, 400. 

Neural networks and, more specifically, the LSTM are used for the prediction. We developed several architectures when time series data is only used for training and several other architectures when the combination of time series and static data are used for the training. The followings are the architectures used for training neural networks using the time series measurements.
\begin{figure}[h]
\centering
\begin{subfigure}{.5\linewidth}
  \centering
  \includegraphics[width=.5\linewidth]{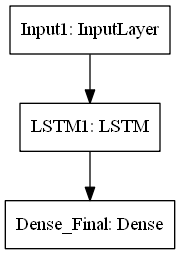}
  \caption{Model 1}
  \label{fig:Img1}
\end{subfigure}%
\begin{subfigure}{.5\linewidth}
  \centering
  \includegraphics[width=.5\linewidth]{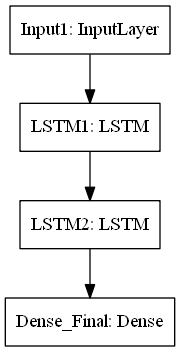}
  \caption{Model2}
  \label{fig:im2}
\end{subfigure}

\begin{subfigure}{.5\textwidth}
  \centering
  \includegraphics[width=.6\linewidth]{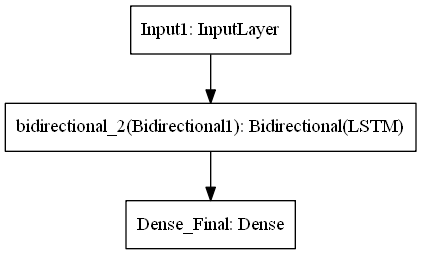}
  \caption{Model3}
  \label{fig:im3}
\end{subfigure}
\caption{Neural Network Architectures for Time Series Data}
\label{fig:Dynamic}
\end{figure}

To combine static and time series data architectures in Fig. 2 are proposed:
\begin{figure}[hbt]
\centering
\begin{subfigure}[b]{.6\linewidth}
  \centering
  \includegraphics[width=.7\linewidth]{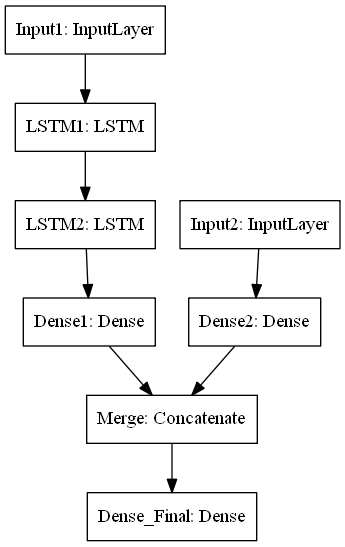}
  \caption{Model 4}
  \label{fig:Img4}
  \vspace{4ex}
\end{subfigure}%
\begin{subfigure}[b]{.6\linewidth}
  \centering
  \includegraphics[width=.7\linewidth]{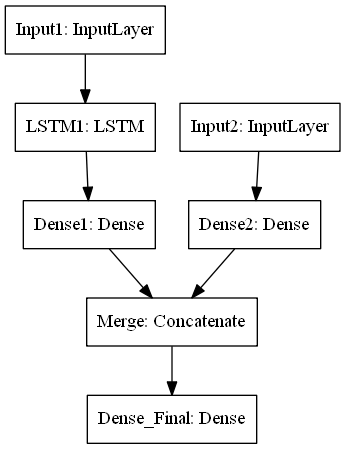}
  \caption{Model5}
  \label{fig:im5}
  \vspace{4ex}
\end{subfigure}
\begin{subfigure}[b]{.45\linewidth}
\centering
  \includegraphics[width=.9\linewidth]{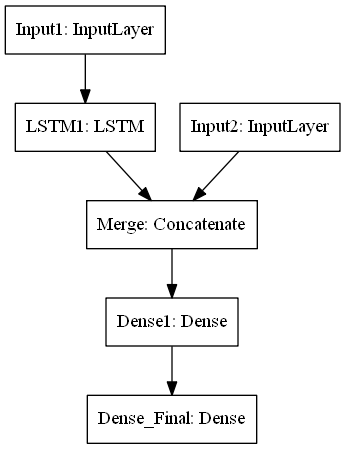}
  \caption{Model6}
  \label{fig:im6}
\end{subfigure}
\hspace{6mm}
\begin{subfigure}[b]{.45\linewidth}
  \centering
  \includegraphics[width=.9\linewidth]{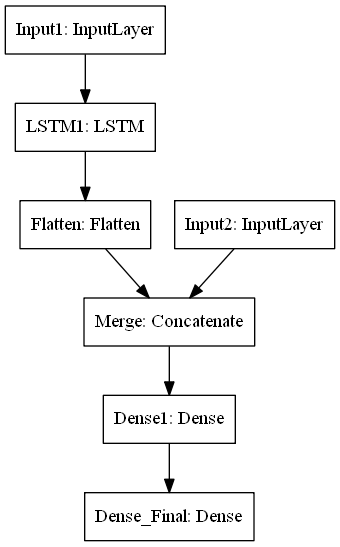}
  \caption{Model7}
  \label{fig:im7}
\end{subfigure}
\caption{Neural Network Architectures for Combining Time Series and Static Data}
\label{fig:Dynamic&Static}
\end{figure}

 Architectures in Fig. \ref{fig:Dynamic} and \ref{fig:Dynamic&Static} are tested against clustered and unclustered data under the following conditions:
 \begin{itemize}
     \item Only gas measurement (third column of time series data) and the static data are used to predict the future values of gas.
     \item All time series measurements or a subset of time series measurements, in addition to static data, are used to predict the future value gas.
 \end{itemize}
 \section{Results and Discussion}
In this section, we describe the results for anomaly detection as well as forecasting with and without clustering. 
The plots in Fig. 3 present some examples of anomaly detection algorithm presented in section II. 
\begin{figure}[ht]
\centering
\begin{subfigure}[b]{.5\linewidth}
  \centering
  \includegraphics[width=1\linewidth]{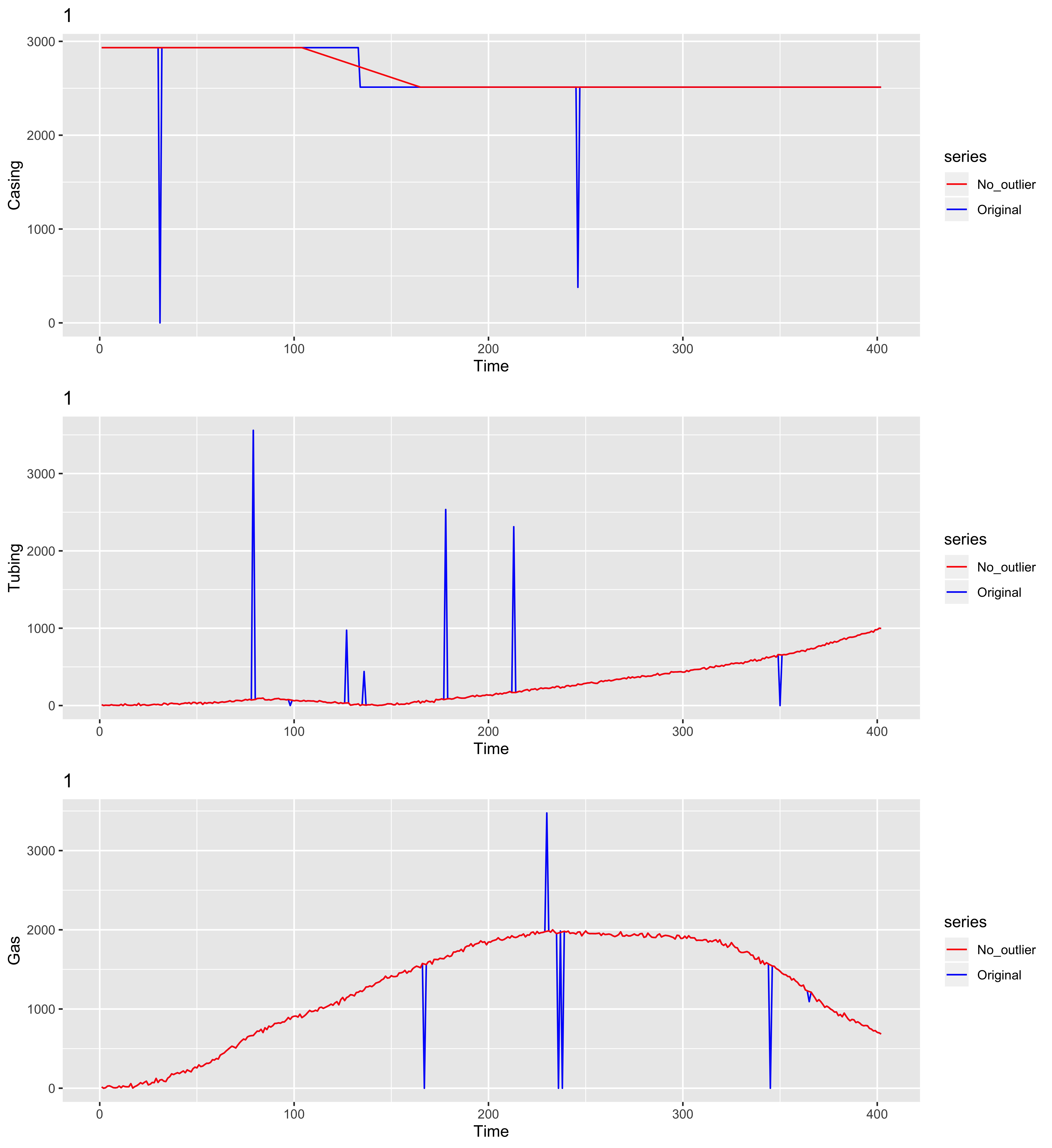}
  \vspace{4ex}
\end{subfigure}%
\begin{subfigure}[b]{.5\linewidth}
  \centering
  \includegraphics[width=1\linewidth]{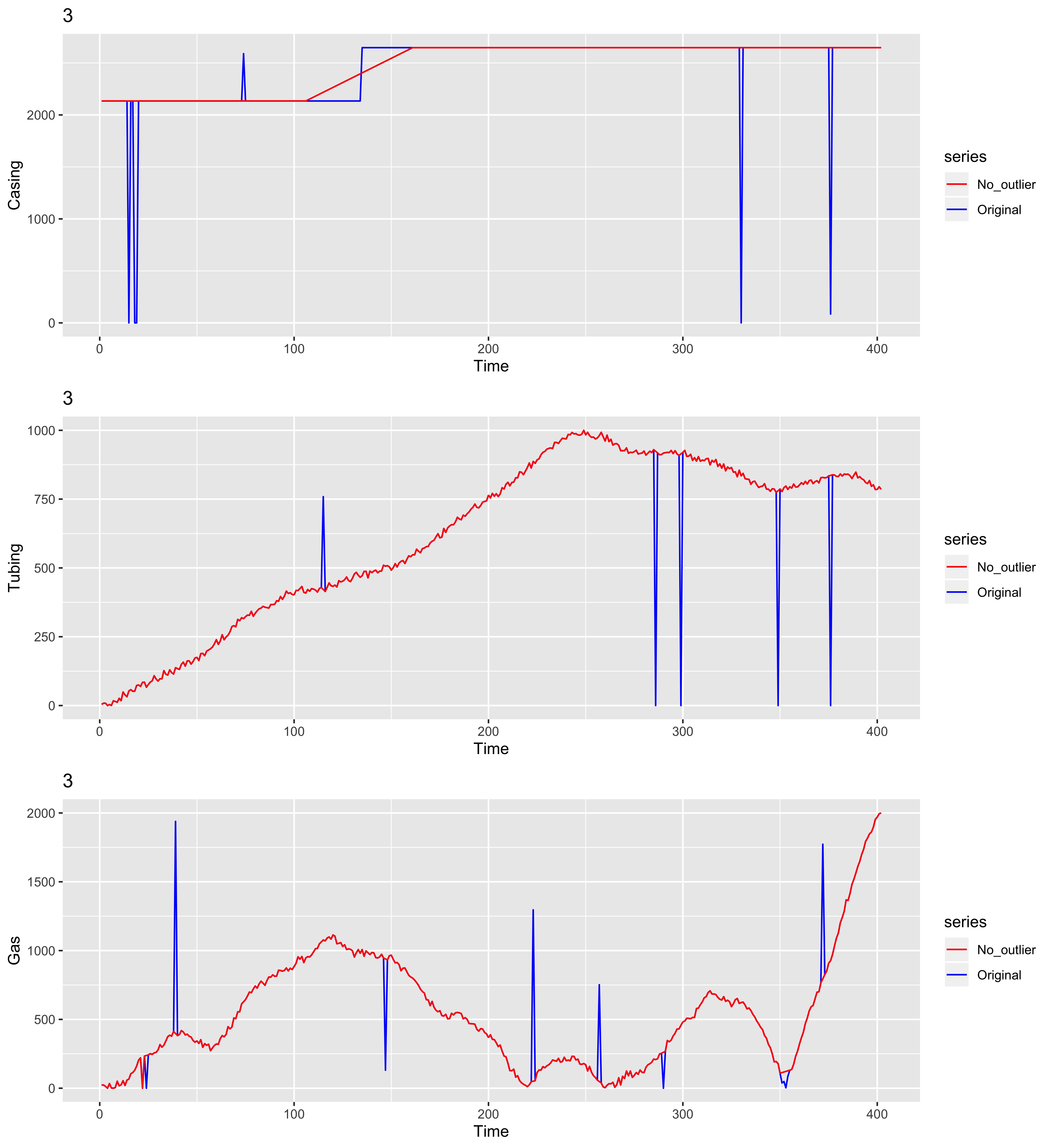}
  \vspace{4ex}
\end{subfigure}

\begin{subfigure}[b]{.6\linewidth}
  \centering
  \includegraphics[width=1\linewidth]{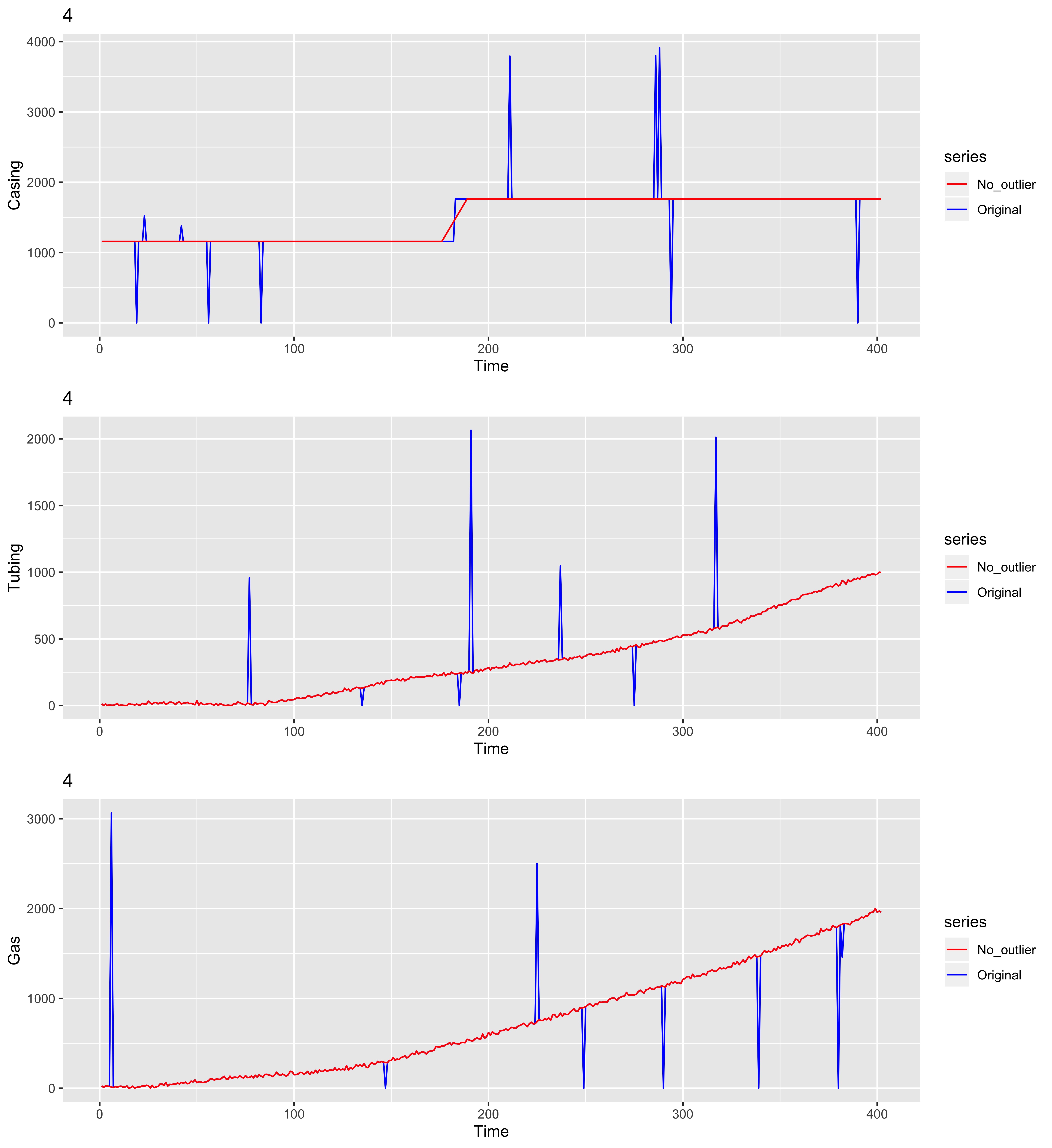}
  \vspace{4ex}
\end{subfigure}
\caption{Anomaly Detection Algorithm}
\label{fig:Anomaly}
\end{figure}
In Fig. \ref{fig:Anomaly} the blue curve is the original curve, and the red is the curve when outliers are removed. As it is shown, the anomaly detection algorithm described in section II can accurately identify the outliers. This is a crucial preprocessing step to make sure the data used for prediction is reliable.

Table III presents the MAPE, MAE, and RMSE error measures using 5-fold cross-validations for different neural network architectures introduced in the last section when only gas measurement (third column of time series) is used to predict the future values of gas. 
\begin{table}[tbh]
\caption{Forecasting Results with No Clustering}
\resizebox{9cm}{!}{
\begin{tabular}{ |c|c|c|c|c|c| } 
\hline
Architecture & Performance Measure &  150 & 200 & 300 & 400 \\
\hline
\multirow{4}{4em}{Model 1} & RMSE & 9684.88 & 27005.89 & 57649.70 & 71668.95 \\
 & MAPE & \textcolor{blue}{33.94} & 42.41 & 44.54 & 20.51 \\ 
& MAE & 6827.62 & 19259.43 & 44179.18 & 53633.80 \\ 
\hline
& Time(s) & 1754 & 1613 & 1420 & 1250\\
\hline
\multirow{4}{4em}{Model 2} & RMSE & 9736.41 & 25782.56 & 53861.46 & 69246.07 \\
 & MAPE & 35.53 & 40.50 & 38.69 & 18.88 \\ 
& MAE & 6716.27 & \textcolor{cyan}{17245.11} & 43024.10 & 54552.27 \\ 
\hline
& Time(s) & 3003 & 2521 & 2067 & 2125\\
\hline
\multirow{4}{4em}{Model 3} & RMSE & \textcolor{red}{9216.22} & 26161.17 & \textcolor{red}{50678.83} & \textcolor{red}{65593.37} \\
 & MAPE & 35.42 & \textcolor{blue}{40.37} & \textcolor{blue}{38.52} & \textcolor{blue}{18.18} \\ 
& MAE & \textcolor{cyan}{6462.38} & 19731.54 & 45336.42 & \textcolor{cyan}{51144.22} \\ 
\hline
& Time(s) & 1900 & 1861 & 1600 & 1550\\
\hline
\multirow{4}{4em}{Model 4} & RMSE & 10866.09 & \textcolor{red}{24287.97} & 56205.88 & 73167.68 \\
 & MAPE & 44.19 & 44.12 & 38.73 & 20.98 \\ 
& MAE & 8065.32 & 17892.22 & 43378.11 & 57337.14 \\ 
\hline
& Time(s) & 2900 & 2851 & 3421 & 2550\\
\hline
\multirow{4}{4em}{Model 5} & RMSE & 10313.77 & 26131.29 & 53851.36 & 72557.48 \\
 & MAPE & 40.17 & 46.08 & 37.63 & 21.07 \\ 
& MAE & 7533.38 & 19163.40 & \textcolor{cyan}{42630.06} & 55765.28 \\ 
\hline
& Time(s) & 1228 & 1888 & 1536 & 1409\\
\hline
\multirow{4}{4em}{Model 6} & RMSE & 11473.68 & 26938.94 & 54924.93 & 74088.37 \\
 & MAPE & 45.24 & 56.77 & 38.54 & 22.62 \\ 
& MAE & 8042.11 & 20872.36 & 44216.48 & 59516.29 \\ 
\hline
& Time(s) & 1800 & 1600 & 1770 & 1662\\
\hline
\multirow{4}{4em}{Model 7} & RMSE & 16198.66 & 32106.55 & 66425.61 & 76797.46 \\
 & MAPE & 71.09 & 71.10 & 41.13 & 22.56 \\ 
& MAE & 12662.59 & 25438.29 & 52732.66 & 61946.76 \\ 
\hline
& Time(s) & 1600 & 1523 & 1327 & 1710\\
\hline
\end{tabular}
}

\end{table}

In Table III, red color identifies the best RMSE; blue identifies best MAPE, and cyan identifies best MAE values among all architectures. According to Table III, Model 3 (Bidirectional LSTM) overall has the best performance since it gives the best RMSE values for K=150, 300, 400, and best MAPE values for K=200, 300, 400. In terms of time, Model 4 has the highest average time complexity with 2938 (s). Prediction without clustering can have the following problems:
\begin{itemize}
    \item Long forecasting time. If the process is interrupted due to insufficient memory or CPU, then training and forecasting should be repeated for all the data again.  
    \item Since data exhibit various patterns, shapes, and characteristics, training all of them together might lower the prediction performance.
\end{itemize}

Table IV summarizes the clustering characteristics for various clustering algorithms described in section II part C.
\begin{table}[hbt]
\caption{Characteristics of Various Clustering Algorithms}
\resizebox{1.1\linewidth}{!}{
\begin{tabular}{cc|c|c|c|c}
\cline{3-5}
& & Number of Measurements & Time(s) & Number of cluster  \\ \cline{1-5}
\multicolumn{1}{ |c  }{\multirow{2}{*}{Distance-Based} } &
\multicolumn{1}{ |c| }{\multirow{2}{*}{DTW} } & 1 & 1352.40 & 3 &      \\ \cline{3-5}
\multicolumn{1}{ |c  }{}                        &
\multicolumn{1}{ |c| }{} & 3 & 1486.57 & 3 &      \\ \cline{1-5}
\multicolumn{1}{ |c  }{\multirow{4}{*}{Feature-Based} } &
\multicolumn{1}{ |c| }{\multirow{2}{*}{Method A} } & 1 & 12.61 & 2 &  \\ \cline{3-5}
\multicolumn{1}{ |c  }{}                            &
\multicolumn{1}{ |c| }{} & 3 & 24.48 & 2 &   \\ \cline{2-5}
\multicolumn{1}{ |c| }{} & \multirow{2}{*}{Method B} & 1 & 275.71 & 2 &  \\ \cline{3-5}
\multicolumn{1}{ |c  }{}                            &
\multicolumn{1}{ |c| }{} & 3 & 406.52 & 2 &  \\ \cline{1-5}

\end{tabular}
}
\end{table}
According to Table IV, feature-based methods outperform the distance-based method in terms of speed; moreover, Method A is faster of the two feature-based methods. The optimal number of clusters in Table IV is determined using several cluster validity indexes (CVI) such as Silhouette, Dunn, and Gamma.
Next, we will present the prediction results with DTW clustering when only the gas measurement is used to predict the future value of the gas. 
\begin{table}
\caption{Forecasting Results with DTW Clustering}
\large
\resizebox{0.5\textwidth}{!}{

\begin{tabular}{ |cccc|c|c|c|c|c|c| } 
\cline{2-7}
\multicolumn{1}{c|}{\multirow{28}{*}{\textbf{Cluster 1}}} & \multicolumn{1}{ |c| } {Architecture} & \multicolumn{1}{ |c| }{Performance Measure} &  150 & 200 & 300 & 400 \\ \cline{1-7}

\multicolumn{1}{ |c| }{} & \multicolumn{1}{ |c| }{\multirow{4}{*}{Model 1}} & \multicolumn{1}{ |c| }{RMSE} & \textcolor{red}{13492.88} & \textcolor{red}{27227.05} & 49960.58 & \textcolor{red}{71390.70}  \\
\multicolumn{1}{ |c| }{} & \multicolumn{1}{ |c| }{} & \multicolumn{1}{ |c| }{MAPE} & \textcolor{blue}{14.18} & 17.73 & 15.95 & \textcolor{blue}{15.44}  \\ 
\multicolumn{1}{ |c| }{} & \multicolumn{1}{ |c| }{} & \multicolumn{1}{ |c| }{MAE} & \textcolor{cyan}{10493.94} & \textcolor{cyan}{19177.90} & 39821.71 & 57437.43 \\ 
\multicolumn{1}{ |c| }{} & \multicolumn{1}{ |c| }{} & \multicolumn{1}{ |c| }{Time(s)} & 378 & 340 & 400 & 391\\
\cline{2-7}
\multicolumn{1}{ |c| }{} & \multicolumn{1}{ |c| }{\multirow{4}{*}{Model 2}} & \multicolumn{1}{ |c| }{RMSE} & 15664.32 & 29477.30 & 50128.86 & 72614.45  \\
\multicolumn{1}{ |c| }{} & \multicolumn{1}{ |c| }{} & \multicolumn{1}{ |c| }{MAPE} & {16.13} & 18.19 & 15.80 & 15.47  \\ 
\multicolumn{1}{ |c| }{} & \multicolumn{1}{ |c| }{} & \multicolumn{1}{ |c| }{MAE} & 13932.04 & 23122.89 & 38783.94 & 60840.61 \\ 
\multicolumn{1}{ |c| }{} & \multicolumn{1}{ |c| }{} & \multicolumn{1}{ |c| }{Time(s)} & 687 & 1044 & 971 & 1200\\
\cline{2-7}
\multicolumn{1}{ |c| }{} & \multicolumn{1}{ |c| }{\multirow{4}{*}{Model 3}} & \multicolumn{1}{ |c| }{RMSE} & 14319.32 & 29365.25 & 50414.84 & 72061.09  \\
\multicolumn{1}{ |c| }{} & \multicolumn{1}{ |c| }{} & \multicolumn{1}{ |c| }{MAPE} & {14.91} & 17.05 & 16.56 & 16.16  \\ 
\multicolumn{1}{ |c| }{} & \multicolumn{1}{ |c| }{} & \multicolumn{1}{ |c| }{MAE} & 10814.49 & 24784.66 & 42159.97 & 60357.10 \\ 
\multicolumn{1}{ |c| }{} & \multicolumn{1}{ |c| }{} & \multicolumn{1}{ |c| }{Time(s)} & 503 & 600 & 726 & 800\\
\cline{2-7}
\multicolumn{1}{ |c| }{} & \multicolumn{1}{ |c| }{\multirow{4}{*}{Model 4}} & \multicolumn{1}{ |c| }{RMSE} & 17292.00 & 32476.94 & \textcolor{red}{44582.48} & 72725.99  \\
\multicolumn{1}{ |c| }{} & \multicolumn{1}{ |c| }{} & \multicolumn{1}{ |c| }{MAPE} & {19.78} & 18.94 & \textcolor{blue}{12.88} & 16.39  \\ 
\multicolumn{1}{ |c| }{} & \multicolumn{1}{ |c| }{} & \multicolumn{1}{ |c| }{MAE} & 14332.13 & 25061.84 & \textcolor{cyan}{38270.20} & 61102.26 \\ 
\multicolumn{1}{ |c| }{} & \multicolumn{1}{ |c| }{} & \multicolumn{1}{ |c| }{Time(s)} & 1000 & 1080 & 1087 & 1200\\
\cline{2-7}
\multicolumn{1}{ |c| }{} & \multicolumn{1}{ |c| }{\multirow{4}{*}{Model 5}} & \multicolumn{1}{ |c| }{RMSE} & 15303.55 & 31857.90 & 61536.90 & 79902.93  \\
\multicolumn{1}{ |c| }{} & \multicolumn{1}{ |c| }{} & \multicolumn{1}{ |c| }{MAPE} & {16.95} & 18.27 & 21.28 & 18.20  \\ 
\multicolumn{1}{ |c| }{} & \multicolumn{1}{ |c| }{} & \multicolumn{1}{ |c| }{MAE} & 12397.09 & 25548.43 & 49572.39 & 66493.49 \\ 
\multicolumn{1}{ |c| }{} & \multicolumn{1}{ |c| }{} & \multicolumn{1}{ |c| }{Time(s)} & 821 & 546 & 612 & 782\\
\cline{2-7}
\multicolumn{1}{ |c| }{} & \multicolumn{1}{ |c| }{\multirow{4}{*}{Model 6}} & \multicolumn{1}{ |c| }{RMSE} & 14399.24 & 32669.23 & 54847.72 & 82953.44  \\
\multicolumn{1}{ |c| }{} & \multicolumn{1}{ |c| }{} & \multicolumn{1}{ |c| }{MAPE} & {14.89} & 17.93 & 17.62 & 19.19  \\ 
\multicolumn{1}{ |c| }{} & \multicolumn{1}{ |c| }{} & \multicolumn{1}{ |c| }{MAE} & 11715.45 & 19259.43 & 44179.18 & \textcolor{cyan}{53633.80} \\ 
\multicolumn{1}{ |c| }{} & \multicolumn{1}{ |c| }{} & \multicolumn{1}{ |c| }{Time(s)} & 453 & 460 & 433 & 421\\
\cline{2-7}
\multicolumn{1}{ |c| }{} & \multicolumn{1}{ |c| }{\multirow{4}{*}{Model 7}} & \multicolumn{1}{ |c| }{RMSE} & 19112.97 & 32594.89 & 51854.97 & 75612.36  \\
\multicolumn{1}{ |c| }{} & \multicolumn{1}{ |c| }{} & \multicolumn{1}{ |c| }{MAPE} & {19.12} & \textcolor{blue}{16.24} & 15.86 & 16.76  \\ 
\multicolumn{1}{ |c| }{} & \multicolumn{1}{ |c| }{} & \multicolumn{1}{ |c| }{MAE} & 15384.47 & 25164.28 & 40750.27 & 65085.91 \\ 
\multicolumn{1}{ |c| }{} & \multicolumn{1}{ |c| }{} & \multicolumn{1}{ |c| }{Time(s)} & 430 & 491 & 455 & 435\\
\cline{1-7}

\multicolumn{1}{c|}{\multirow{28}{*}{\textbf{Cluster 2}}} & \multicolumn{1}{ |c| } {Architecture} & \multicolumn{1}{ |c| }{Performance Measure} &  150 & 200 & 300 & 400 \\ \cline{1-7}

\multicolumn{1}{ |c| }{} & \multicolumn{1}{ |c| }{\multirow{4}{*}{Model 1}} & \multicolumn{1}{ |c| }{RMSE} & 4414.00 & 10923.24 & 31035.56 & 52708.65  \\
\multicolumn{1}{ |c| }{} & \multicolumn{1}{ |c| }{} & \multicolumn{1}{ |c| }{MAPE} & {33.37} & 43.41 & 40.05 & 20.52 \\ 
\multicolumn{1}{ |c| }{} & \multicolumn{1}{ |c| }{} & \multicolumn{1}{ |c| }{MAE} & 3640.56 & 9157.58 & 25500.18 & 41605.72 \\ 
\multicolumn{1}{ |c| }{} & \multicolumn{1}{ |c| }{} & \multicolumn{1}{ |c| }{Time(s)} & 800 & 557 & 800 & 871\\
\cline{2-7}
\multicolumn{1}{ |c| }{} & \multicolumn{1}{ |c| }{\multirow{4}{*}{Model 2}} & \multicolumn{1}{ |c| }{RMSE} & 4419.03 & 11557.68 & 31079.12 & 49957.10  \\
\multicolumn{1}{ |c| }{} & \multicolumn{1}{ |c| }{} & \multicolumn{1}{ |c| }{MAPE} & {29.67} & 46.51 & 38.68 & 19.26  \\ 
\multicolumn{1}{ |c| }{} & \multicolumn{1}{ |c| }{} & \multicolumn{1}{ |c| }{MAE} & 3452.11 & 9547.25 & 25161.26 & 39200.76 \\ 
\multicolumn{1}{ |c| }{} & \multicolumn{1}{ |c| }{} & \multicolumn{1}{ |c| }{Time(s)} & 1450 & 1200 & 1745 & 1800\\
\cline{2-7}
\multicolumn{1}{ |c| }{} & \multicolumn{1}{ |c| }{\multirow{4}{*}{Model 3}} & \multicolumn{1}{ |c| }{RMSE} & 4375.45 & 10533.33 & \textcolor{red}{30225.99} & \textcolor{red}{47035.53}  \\
\multicolumn{1}{ |c| }{} & \multicolumn{1}{ |c| }{} & \multicolumn{1}{ |c| }{MAPE} & {32.53} & \textcolor{blue}{43.11} & \textcolor{blue}{38.26} & \textcolor{blue}{16.62}  \\ 
\multicolumn{1}{ |c| }{} & \multicolumn{1}{ |c| }{} & \multicolumn{1}{ |c| }{MAE} & 3827.21 & 9393.74 & \textcolor{cyan}{24633.34} & \textcolor{cyan}{37753.89} \\ 
\multicolumn{1}{ |c| }{} & \multicolumn{1}{ |c| }{} & \multicolumn{1}{ |c| }{Time(s)} & 952 & 942 & 923 & 968\\
\cline{2-7}
\multicolumn{1}{ |c| }{} & \multicolumn{1}{ |c| }{\multirow{4}{*}{Model 4}} & \multicolumn{1}{ |c| }{RMSE} & \textcolor{red}{4028.32} & \textcolor{red}{10457.87} & 34559.18 & 57166.64  \\
\multicolumn{1}{ |c| }{} & \multicolumn{1}{ |c| }{} & \multicolumn{1}{ |c| }{MAPE} & \textcolor{blue}{27.26} & 45.30 & 47.08 & 22.70  \\ 
\multicolumn{1}{ |c| }{} & \multicolumn{1}{ |c| }{} & \multicolumn{1}{ |c| }{MAE} & \textcolor{cyan}{3297.35} & \textcolor{cyan}{8413.90} & 29013.81 & 44908.45 \\ 
\multicolumn{1}{ |c| }{} & \multicolumn{1}{ |c| }{} & \multicolumn{1}{ |c| }{Time(s)} & 1379 & 1800 & 1826 & 1913\\
\cline{2-7}
\multicolumn{1}{ |c| }{} & \multicolumn{1}{ |c| }{\multirow{4}{*}{Model 5}} & \multicolumn{1}{ |c| }{RMSE} & 4604.66 & 12290.53 & 33935.11 & 53214.73  \\
\multicolumn{1}{ |c| }{} & \multicolumn{1}{ |c| }{} & \multicolumn{1}{ |c| }{MAPE} & {31.24} & 52.95 & 41.64 & 20.69  \\ 
\multicolumn{1}{ |c| }{} & \multicolumn{1}{ |c| }{} & \multicolumn{1}{ |c| }{MAE} & 3623.62 & 10039.21 & 27668.60 & 42097.45 \\ 
\multicolumn{1}{ |c| }{} & \multicolumn{1}{ |c| }{} & \multicolumn{1}{ |c| }{Time(s)} & 584 & 713 & 920 & 678\\
\cline{2-7}
\multicolumn{1}{ |c| }{} & \multicolumn{1}{ |c| }{\multirow{4}{*}{Model 6}} & \multicolumn{1}{ |c| }{RMSE} & 4413.09 & 10595.32 & 33016.45 & 54944.10  \\
\multicolumn{1}{ |c| }{} & \multicolumn{1}{ |c| }{} & \multicolumn{1}{ |c| }{MAPE} & {33.25} & 44.65 & 39.54 & 19.40  \\ 
\multicolumn{1}{ |c| }{} & \multicolumn{1}{ |c| }{} & \multicolumn{1}{ |c| }{MAE} & 3562.62 & 8526.28 & 27270.80 & 42646.56 \\ 
\multicolumn{1}{ |c| }{} & \multicolumn{1}{ |c| }{} & \multicolumn{1}{ |c| }{Time(s)} & 715 & 880 & 979 & 917\\
\cline{2-7}
\multicolumn{1}{ |c| }{} & \multicolumn{1}{ |c| }{\multirow{4}{*}{Model 7}} & \multicolumn{1}{ |c| }{RMSE} & 5012.49 & 12623.80 & 33588.65 & 51467.56  \\
\multicolumn{1}{ |c| }{} & \multicolumn{1}{ |c| }{} & \multicolumn{1}{ |c| }{MAPE} & {41.53} & 53.82 & 42.48 & 19.79 \\ 
\multicolumn{1}{ |c| }{} & \multicolumn{1}{ |c| }{} & \multicolumn{1}{ |c| }{MAE} & 4225.53 & 10208.43 & 27638.36 & 41616.74 \\ 
\multicolumn{1}{ |c| }{} & \multicolumn{1}{ |c| }{} & \multicolumn{1}{ |c| }{Time(s)} & 712 & 725 & 781 & 796\\
\cline{1-7}

\multicolumn{1}{c|}{\multirow{28}{*}{\textbf{Cluster 3}}} & \multicolumn{1}{ |c| } {Architecture} & \multicolumn{1}{ |c| }{Performance Measure} &  150 & 200 & 300 & 400 \\ \cline{1-7}

\multicolumn{1}{ |c| }{} & \multicolumn{1}{ |c| }{\multirow{4}{*}{Model 1}} & \multicolumn{1}{ |c| }{RMSE} & 8035.00 & 20863.14 & 41267.41 & \textcolor{red}{50736.26}  \\
\multicolumn{1}{ |c| }{} & \multicolumn{1}{ |c| }{} & \multicolumn{1}{ |c| }{MAPE} & {14.91} & 19.95 & 20.11 & 12.27  \\ 
\multicolumn{1}{ |c| }{} & \multicolumn{1}{ |c| }{} & \multicolumn{1}{ |c| }{MAE} & 5999.42 & 16170.73 & 34492.70 & \textcolor{cyan}{41323.80} \\ 
\multicolumn{1}{ |c| }{} & \multicolumn{1}{ |c| }{} & \multicolumn{1}{ |c| }{Time(s)} & 542 & 646 & 632 & 753\\
\cline{2-7}
\multicolumn{1}{ |c| }{} & \multicolumn{1}{ |c| }{\multirow{4}{*}{Model 2}} & \multicolumn{1}{ |c| }{RMSE} & 8138.11 & 18217.21 & 45578.03 & 55843.99  \\
\multicolumn{1}{ |c| }{} & \multicolumn{1}{ |c| }{} & \multicolumn{1}{ |c| }{MAPE} & \textcolor{blue}{14.35} & \textcolor{blue}{17.24} & 19.37 & \textcolor{blue}{12.15}  \\ 
\multicolumn{1}{ |c| }{} & \multicolumn{1}{ |c| }{} & \multicolumn{1}{ |c| }{MAE} & \textcolor{cyan}{5702.11} & 12931.99 & 34985.26 & 44785.75 \\ 
\multicolumn{1}{ |c| }{} & \multicolumn{1}{ |c| }{} & \multicolumn{1}{ |c| }{Time(s)} & 1020 & 879 & 1225 & 1410\\
\cline{2-7}
\multicolumn{1}{ |c| }{} & \multicolumn{1}{ |c| }{\multirow{4}{*}{Model 3}} & \multicolumn{1}{ |c| }{RMSE} & 9606.88 & 19171.64 & 50842.20 & 65310.41  \\
\multicolumn{1}{ |c| }{} & \multicolumn{1}{ |c| }{} & \multicolumn{1}{ |c| }{MAPE} & {17.07} & 20.13 & 19.33 & 17.51  \\ 
\multicolumn{1}{ |c| }{} & \multicolumn{1}{ |c| }{} & \multicolumn{1}{ |c| }{MAE} & 6507.99 & 14906.23 & 36474.90 & 51834.14 \\ 
\multicolumn{1}{ |c| }{} & \multicolumn{1}{ |c| }{} & \multicolumn{1}{ |c| }{Time(s)} & 802 & 1007 & 921 & 879\\
\cline{2-7}
\multicolumn{1}{ |c| }{} & \multicolumn{1}{ |c| }{\multirow{4}{*}{Model 4}} & \multicolumn{1}{ |c| }{RMSE} & 9513.50 & 24396.60 & 36394.98 & 62606.17  \\
\multicolumn{1}{ |c| }{} & \multicolumn{1}{ |c| }{} & \multicolumn{1}{ |c| }{MAPE} & {19.66} & 19.87 & 17.92 & 15.15  \\ 
\multicolumn{1}{ |c| }{} & \multicolumn{1}{ |c| }{} & \multicolumn{1}{ |c| }{MAE} & 7497.21 & 16790.39 & 29828.65 & 48750.67 \\ 
\multicolumn{1}{ |c| }{} & \multicolumn{1}{ |c| }{} & \multicolumn{1}{ |c| }{Time(s)} & 1414 & 1834 & 1821 & 1690\\
\cline{2-7}
\multicolumn{1}{ |c| }{} & \multicolumn{1}{ |c| }{\multirow{4}{*}{Model 5}} & \multicolumn{1}{ |c| }{RMSE} & 8857.02 & 22665.07 & \textcolor{red}{35642.91} & 59074.92  \\
\multicolumn{1}{ |c| }{} & \multicolumn{1}{ |c| }{} & \multicolumn{1}{ |c| }{MAPE} & {17.31} & 18.82 & \textcolor{blue}{16.41} & 14.48  \\ 
\multicolumn{1}{ |c| }{} & \multicolumn{1}{ |c| }{} & \multicolumn{1}{ |c| }{MAE} & 6841.79 & 16854.21 & \textcolor{cyan}{28238.08} & 47178.45 \\ 
\multicolumn{1}{ |c| }{} & \multicolumn{1}{ |c| }{} & \multicolumn{1}{ |c| }{Time(s)} & 973 & 767 & 718 & 723\\
\cline{2-7}
\multicolumn{1}{ |c| }{} & \multicolumn{1}{ |c| }{\multirow{4}{*}{Model 6}} & \multicolumn{1}{ |c| }{RMSE} & \textcolor{red}{7891.00} & \textcolor{red}{16401.70} & 38307.10 & 60533.33  \\
\multicolumn{1}{ |c| }{} & \multicolumn{1}{ |c| }{} & \multicolumn{1}{ |c| }{MAPE} & 15.09 & 17.90 & 18.70 & 15.72 \\    
\multicolumn{1}{ |c| }{} & \multicolumn{1}{ |c| }{} & \multicolumn{1}{ |c| }{MAE} & 6014.56 & \textcolor{cyan}{12895.91} & 31065.94 & 45152.46 \\ 
\multicolumn{1}{ |c| }{} & \multicolumn{1}{ |c| }{} & \multicolumn{1}{ |c| }{Time(s)} & 800 & 617 & 648 & 712\\
\cline{2-7}
\multicolumn{1}{ |c| }{} & \multicolumn{1}{ |c| }{\multirow{4}{*}{Model 7}} & \multicolumn{1}{ |c| }{RMSE} & 9100.24 & 17191.88 & 39321.63 & 58002.70  \\
\multicolumn{1}{ |c| }{} & \multicolumn{1}{ |c| }{} & \multicolumn{1}{ |c| }{MAPE} & {19.70} & 19.57 & 19.51 & 13.52  \\ 
\multicolumn{1}{ |c| }{} & \multicolumn{1}{ |c| }{} & \multicolumn{1}{ |c| }{MAE} & 7582.63 & 13583.04 & 32742.17 & 44045.32 \\ 
\multicolumn{1}{ |c| }{} & \multicolumn{1}{ |c| }{} & \multicolumn{1}{ |c| }{Time(s)} & 723 & 711 & 689 & 656\\
\cline{1-7}
\end{tabular}
}

\end{table}

According to Table V clustering has the following advantages:
\begin{itemize}
    \item Prediction for different clusters can be made at different times, or it can be done parallel, which significantly reduces the forecasting time. 
    \item For each K, the best architecture for different clusters can be chosen to improve the overall performance. For example, consider K=150 when there is no clustering; the error metrics are as follows: RMSE =9. 216, MAPE = 33.94 and MAE = 6462.38. Now when DTW clustering is used for prediction given that 94 time series are in cluster 1, 188 time series are in cluster 2 and 118 time series are in cluster 3 the error metrics are calculated as follows:
    \begin{align}
         E_{Tot}  = \Sigma_{i} P_{i} E_{i} 
    \end{align}
    \begin{equation}
    \begin{split}
         RMSE_{Tot}  = \dfrac{94}{400} \times 13492.88 + \dfrac{188}{400}\\ \times 4028 + \dfrac{118}{400} \times 7891 = 7391.83
         \end{split}
    \end{equation}
    \begin{equation}
    \begin{split}
         MAPE_{Tot}  = \dfrac{94}{400} \times 14.18 + \dfrac{188}{400}\\ \times 27.26 + \dfrac{118}{400} \times 14.35 = 20.38
         \end{split}
    \end{equation}
    \begin{equation}
    \begin{split}
         MAE_{Tot}  = \dfrac{94}{400} \times 10493.94 + \dfrac{188}{400}\\ \times 3297.35 + \dfrac{118}{400} \times 5702.11 = 5697.73
         \end{split}
    \end{equation}
    In (6) $E_{Tot}$ is the total error, $E_{i}$ is the error corresponding to each cluster and $P_{i}$ is the probability of choosing a cluster. Equation (6) follows the law of total probability. 
\end{itemize}
Comparing the results in (7), (8) and (9) with RMSE, MAPE, and MAE in Table III reveal that clustering can significantly improve the prediction performance. 

Tables VI and VII present the forecasting results with feature-based clustering using method A and method B, respectively. 
\begin{table}
\caption{Forecasting Results with Feature-Based Method A Clustering}
\large
\resizebox{0.5\textwidth}{!}{
\begin{tabular}{ |cccc|c|c|c|c|c|c| } 
\cline{2-7}
\multicolumn{1}{c|}{\multirow{28}{*}{\textbf{Cluster 1}}} & \multicolumn{1}{ |c| } {Architecture} & \multicolumn{1}{ |c| }{Performance Measure} &  150 & 200 & 300 & 400 \\ \cline{1-7}

\multicolumn{1}{ |c| }{} & \multicolumn{1}{ |c| }{\multirow{4}{*}{Model 1}} & \multicolumn{1}{ |c| }{RMSE} & {4038.76} & \textcolor{red}{8832.14} & 27142.08 & {45872.40}  \\
\multicolumn{1}{ |c| }{} & \multicolumn{1}{ |c| }{} & \multicolumn{1}{ |c| }{MAPE} & {34.56} & \textcolor{blue}{38.37} & 38.89 & {18.52}  \\ 
\multicolumn{1}{ |c| }{} & \multicolumn{1}{ |c| }{} & \multicolumn{1}{ |c| }{MAE} & \textcolor{cyan}{3204} & \textcolor{cyan}{7341.46} & 23202.30 & 37073.63 \\ 
\multicolumn{1}{ |c| }{} & \multicolumn{1}{ |c| }{} & \multicolumn{1}{ |c| }{Time(s)} & 621 & 691 & 630 & 681\\
\cline{2-7}
\multicolumn{1}{ |c| }{} & \multicolumn{1}{ |c| }{\multirow{4}{*}{Model 2}} & \multicolumn{1}{ |c| }{RMSE} & 4545.92 & 9660.00 & 27239.60 & 44317.91  \\
\multicolumn{1}{ |c| }{} & \multicolumn{1}{ |c| }{} & \multicolumn{1}{ |c| }{MAPE} & {32.44} & 41.28 & 36.48 & 17.11  \\ 
\multicolumn{1}{ |c| }{} & \multicolumn{1}{ |c| }{} & \multicolumn{1}{ |c| }{MAE} & 3400.99 & 7842.32 & 23022.10 & 35806.13 \\ 
\multicolumn{1}{ |c| }{} & \multicolumn{1}{ |c| }{} & \multicolumn{1}{ |c| }{Time(s)} & 1249 & 1398 & 1100 & 1080\\
\cline{2-7}
\multicolumn{1}{ |c| }{} & \multicolumn{1}{ |c| }{\multirow{4}{*}{Model 3}} & \multicolumn{1}{ |c| }{RMSE} & \textcolor{red}{3975.74} & 9906.88 & 27657.59 & \textcolor{red}{42505.00}  \\
\multicolumn{1}{ |c| }{} & \multicolumn{1}{ |c| }{} & \multicolumn{1}{ |c| }{MAPE} & {33.62} & 46.85 & 40.81 & 17.44  \\ 
\multicolumn{1}{ |c| }{} & \multicolumn{1}{ |c| }{} & \multicolumn{1}{ |c| }{MAE} & 3306.95 & 8172.18 & 23897.82 & 35207.36 \\ 
\multicolumn{1}{ |c| }{} & \multicolumn{1}{ |c| }{} & \multicolumn{1}{ |c| }{Time(s)} & 900 & 861 & 812 & 871\\
\cline{2-7}
\multicolumn{1}{ |c| }{} & \multicolumn{1}{ |c| }{\multirow{4}{*}{Model 4}} & \multicolumn{1}{ |c| }{RMSE} & 4328.14 & 10353.43 & {28500.70} & 42600.88  \\
\multicolumn{1}{ |c| }{} & \multicolumn{1}{ |c| }{} & \multicolumn{1}{ |c| }{MAPE} & {36.20} & 49.93 & {37.78} & \textcolor{blue}{15.72}  \\ 
\multicolumn{1}{ |c| }{} & \multicolumn{1}{ |c| }{} & \multicolumn{1}{ |c| }{MAE} & 3472.24 & 8346.69 & {24439.28} & \textcolor{cyan}{34096.96} \\ 
\multicolumn{1}{ |c| }{} & \multicolumn{1}{ |c| }{} & \multicolumn{1}{ |c| }{Time(s)} & 1313 & 1814 & 1600 & 1324\\
\cline{2-7}
\multicolumn{1}{ |c| }{} & \multicolumn{1}{ |c| }{\multirow{4}{*}{Model 5}} & \multicolumn{1}{ |c| }{RMSE} & 4949.87 & 9389.69 & \textcolor{red}{25986.47} & 45082.67  \\
\multicolumn{1}{ |c| }{} & \multicolumn{1}{ |c| }{} & \multicolumn{1}{ |c| }{MAPE} & {42.09} & 38.83 & \textcolor{blue}{33.53} & 18.50  \\ 
\multicolumn{1}{ |c| }{} & \multicolumn{1}{ |c| }{} & \multicolumn{1}{ |c| }{MAE} & 3951.77 & 7696.16 & \textcolor{cyan}{21675.31} & 36561.82 \\ 
\multicolumn{1}{ |c| }{} & \multicolumn{1}{ |c| }{} & \multicolumn{1}{ |c| }{Time(s)} & 671 & 671 & 712 & 650\\
\cline{2-7}
\multicolumn{1}{ |c| }{} & \multicolumn{1}{ |c| }{\multirow{4}{*}{Model 6}} & \multicolumn{1}{ |c| }{RMSE} & 4279.94 & 9743.54 & 27923.25 & 45612.06  \\
\multicolumn{1}{ |c| }{} & \multicolumn{1}{ |c| }{} & \multicolumn{1}{ |c| }{MAPE} & \textcolor{blue}{32.34} & 54.16 & 40.99 & 17.29  \\ 
\multicolumn{1}{ |c| }{} & \multicolumn{1}{ |c| }{} & \multicolumn{1}{ |c| }{MAE} & 3413.10 & 8544.21 & 23223.17 & {36221.62} \\ 
\multicolumn{1}{ |c| }{} & \multicolumn{1}{ |c| }{} & \multicolumn{1}{ |c| }{Time(s)} & 591 & 660 & 687 & 612\\
\cline{2-7}
\multicolumn{1}{ |c| }{} & \multicolumn{1}{ |c| }{\multirow{4}{*}{Model 7}} & \multicolumn{1}{ |c| }{RMSE} & {4972.13} & {10147.80} & 27145.73 & {46906.73}  \\
\multicolumn{1}{ |c| }{} & \multicolumn{1}{ |c| }{} & \multicolumn{1}{ |c| }{MAPE} & {38.52} & {45.45} & 38.78 & 18.25  \\ 
\multicolumn{1}{ |c| }{} & \multicolumn{1}{ |c| }{} & \multicolumn{1}{ |c| }{MAE} & 4040.88 & 8444.79 & 22175.59 & 38482.42 \\ 
\multicolumn{1}{ |c| }{} & \multicolumn{1}{ |c| }{} & \multicolumn{1}{ |c| }{Time(s)} & 623 & 711 & 601 & 589\\
\cline{1-7}

\multicolumn{1}{c|}{\multirow{28}{*}{\textbf{Cluster 2}}} & \multicolumn{1}{ |c| } {Architecture} & \multicolumn{1}{ |c| }{Performance Measure} &  150 & 200 & 300 & 400 \\ \cline{1-7}

\multicolumn{1}{ |c| }{} & \multicolumn{1}{ |c| }{\multirow{4}{*}{Model 1}} & \multicolumn{1}{ |c| }{RMSE} & \textcolor{red}{11872.12} & 29413.02 & 56085.34 & 65408.21  \\
\multicolumn{1}{ |c| }{} & \multicolumn{1}{ |c| }{} & \multicolumn{1}{ |c| }{MAPE} & {28.97} & 29.99 & 25.27 & 15.75 \\ 
\multicolumn{1}{ |c| }{} & \multicolumn{1}{ |c| }{} & \multicolumn{1}{ |c| }{MAE} & \textcolor{cyan}{8838.86} & \textcolor{cyan}{22715.02} & 46822.08 & 51364.60 \\ 
\multicolumn{1}{ |c| }{} & \multicolumn{1}{ |c| }{} & \multicolumn{1}{ |c| }{Time(s)} & 800 & 882 & 1262 & 922\\
\cline{2-7}
\multicolumn{1}{ |c| }{} & \multicolumn{1}{ |c| }{\multirow{4}{*}{Model 2}} & \multicolumn{1}{ |c| }{RMSE} & 14263.18 & 29695.96 & 53459.04 & 60645.00  \\
\multicolumn{1}{ |c| }{} & \multicolumn{1}{ |c| }{} & \multicolumn{1}{ |c| }{MAPE} & {33.67} & 34.36 & 23.32 & 13.93  \\ 
\multicolumn{1}{ |c| }{} & \multicolumn{1}{ |c| }{} & \multicolumn{1}{ |c| }{MAE} & 11584.29 & 24174.64 & 43544.11 & 47841.09 \\ 
\multicolumn{1}{ |c| }{} & \multicolumn{1}{ |c| }{} & \multicolumn{1}{ |c| }{Time(s)} & 2137 & 1945 & 1462 & 1541\\
\cline{2-7}
\multicolumn{1}{ |c| }{} & \multicolumn{1}{ |c| }{\multirow{4}{*}{Model 3}} & \multicolumn{1}{ |c| }{RMSE} & 14049.78 & 29075.38 & {56893.44} & \textcolor{red}{58575.78}  \\
\multicolumn{1}{ |c| }{} & \multicolumn{1}{ |c| }{} & \multicolumn{1}{ |c| }{MAPE} & {35.24} & {29.36} & {25.76} & \textcolor{blue}{13.61}  \\ 
\multicolumn{1}{ |c| }{} & \multicolumn{1}{ |c| }{} & \multicolumn{1}{ |c| }{MAE} & 11626.64 & 23097.41 & {47130.03} & \textcolor{cyan}{47059.17} \\ 
\multicolumn{1}{ |c| }{} & \multicolumn{1}{ |c| }{} & \multicolumn{1}{ |c| }{Time(s)} & 1162 & 1101 & 1053 & 996\\
\cline{2-7}
\multicolumn{1}{ |c| }{} & \multicolumn{1}{ |c| }{\multirow{4}{*}{Model 4}} & \multicolumn{1}{ |c| }{RMSE} & {15978.33} & {32114.92} & 58155.73 & 63148.46  \\
\multicolumn{1}{ |c| }{} & \multicolumn{1}{ |c| }{} & \multicolumn{1}{ |c| }{MAPE} & {41.68} & 36.38 & 25.97 & 14.26  \\ 
\multicolumn{1}{ |c| }{} & \multicolumn{1}{ |c| }{} & \multicolumn{1}{ |c| }{MAE} & {13298.98} & {26018.09} & 47476.40 & 48745.07 \\ 
\multicolumn{1}{ |c| }{} & \multicolumn{1}{ |c| }{} & \multicolumn{1}{ |c| }{Time(s)} & 1902 & 2084 & 1969 & 2010\\
\cline{2-7}
\multicolumn{1}{ |c| }{} & \multicolumn{1}{ |c| }{\multirow{4}{*}{Model 5}} & \multicolumn{1}{ |c| }{RMSE} & 12947.65 & 29301.81 & 59024.33 & 65776.89  \\
\multicolumn{1}{ |c| }{} & \multicolumn{1}{ |c| }{} & \multicolumn{1}{ |c| }{MAPE} & {30.35} & \textcolor{blue}{29.24} & 25.84 & 16.16  \\ 
\multicolumn{1}{ |c| }{} & \multicolumn{1}{ |c| }{} & \multicolumn{1}{ |c| }{MAE} & 10555.66 & 23042.84 & 46318.66 & 53606.05 \\ 
\multicolumn{1}{ |c| }{} & \multicolumn{1}{ |c| }{} & \multicolumn{1}{ |c| }{Time(s)} & 1239 & 1300 & 1312 & 1080\\
\cline{2-7}
\multicolumn{1}{ |c| }{} & \multicolumn{1}{ |c| }{\multirow{4}{*}{Model 6}} & \multicolumn{1}{ |c| }{RMSE} & 11963.81 & \textcolor{red}{28878.83} & \textcolor{red}{53073.28} & 61825.07  \\
\multicolumn{1}{ |c| }{} & \multicolumn{1}{ |c| }{} & \multicolumn{1}{ |c| }{MAPE} & \textcolor{blue}{26.55} & 33.24 & \textcolor{blue}{23.08} & 14.99  \\ 
\multicolumn{1}{ |c| }{} & \multicolumn{1}{ |c| }{} & \multicolumn{1}{ |c| }{MAE} & 9487.32 & 23477.52 & \textcolor{cyan}{42112.35} & 48812.02 \\ 
\multicolumn{1}{ |c| }{} & \multicolumn{1}{ |c| }{} & \multicolumn{1}{ |c| }{Time(s)} & 1001 & 1211 & 1194 & 951\\
\cline{2-7}
\multicolumn{1}{ |c| }{} & \multicolumn{1}{ |c| }{\multirow{4}{*}{Model 7}} & \multicolumn{1}{ |c| }{RMSE} & 11988.71 & 37235.03 & 61843.47 & 64530.66 \\
\multicolumn{1}{ |c| }{} & \multicolumn{1}{ |c| }{} & \multicolumn{1}{ |c| }{MAPE} & {50.88} & 50.18 & 32.29 & 16.38 \\ 
\multicolumn{1}{ |c| }{} & \multicolumn{1}{ |c| }{} & \multicolumn{1}{ |c| }{MAE} & 9263.22 & 32333.89 & 57271.93 & 53606.42 \\ 
\multicolumn{1}{ |c| }{} & \multicolumn{1}{ |c| }{} & \multicolumn{1}{ |c| }{Time(s)} & 923 & 1187 & 1015 & 986\\
\cline{1-7}
\end{tabular}
}

\end{table}

\begin{table}
\caption{Forecasting Results with Feature-Based Method B Clustering}
\large
\resizebox{0.5\textwidth}{!}{
\begin{tabular}{ |cccc|c|c|c|c|c|c| } 
\cline{2-7}
\multicolumn{1}{c|}{\multirow{28}{*}{\textbf{Cluster 1}}} & \multicolumn{1}{ |c| } {Architecture} & \multicolumn{1}{ |c| }{Performance Measure} &  150 & 200 & 300 & 400 \\ \cline{1-7}

\multicolumn{1}{ |c| }{} & \multicolumn{1}{ |c| }{\multirow{4}{*}{Model 1}} & \multicolumn{1}{ |c| }{RMSE} & {6262.29} & {14602.69} & 40447.69 & {55142.00}  \\
\multicolumn{1}{ |c| }{} & \multicolumn{1}{ |c| }{} & \multicolumn{1}{ |c| }{MAPE} & {34.51} & {43.28} & 48.18 & {19.65}  \\ 
\multicolumn{1}{ |c| }{} & \multicolumn{1}{ |c| }{} & \multicolumn{1}{ |c| }{MAE} & {4705.40} & {11756.55} & 32875.25 & 43141.80 \\ 
\multicolumn{1}{ |c| }{} & \multicolumn{1}{ |c| }{} & \multicolumn{1}{ |c| }{Time(s)} & 1211 & 1031 & 876 & 1105\\
\cline{2-7}
\multicolumn{1}{ |c| }{} & \multicolumn{1}{ |c| }{\multirow{4}{*}{Model 2}} & \multicolumn{1}{ |c| }{RMSE} & 5790.1 & 12494.61 & 33793.05 & 52679.51  \\
\multicolumn{1}{ |c| }{} & \multicolumn{1}{ |c| }{} & \multicolumn{1}{ |c| }{MAPE} & {32.10} & 43.72 & 42.01 & 19.88  \\ 
\multicolumn{1}{ |c| }{} & \multicolumn{1}{ |c| }{} & \multicolumn{1}{ |c| }{MAE} & 4670.78 & 10739.64 & 27789.09 & 42660.21 \\ 
\multicolumn{1}{ |c| }{} & \multicolumn{1}{ |c| }{} & \multicolumn{1}{ |c| }{Time(s)} & 2086 & 2055 & 2010 & 1865\\
\cline{2-7}
\multicolumn{1}{ |c| }{} & \multicolumn{1}{ |c| }{\multirow{4}{*}{Model 3}} & \multicolumn{1}{ |c| }{RMSE} & \textcolor{red}{4906.85} & \textcolor{red}{12144.00} & 32997.05 & \textcolor{red}{49625.67}  \\
\multicolumn{1}{ |c| }{} & \multicolumn{1}{ |c| }{} & \multicolumn{1}{ |c| }{MAPE} & \textcolor{blue}{31.89} & \textcolor{blue}{42.44} & 37.96 & \textcolor{blue}{17.46}  \\ 
\multicolumn{1}{ |c| }{} & \multicolumn{1}{ |c| }{} & \multicolumn{1}{ |c| }{MAE} & \textcolor{cyan}{3591.33} & \textcolor{cyan}{10180.60} & 27072.80 & \textcolor{cyan}{40025.59} \\ 
\multicolumn{1}{ |c| }{} & \multicolumn{1}{ |c| }{} & \multicolumn{1}{ |c| }{Time(s)} & 964 & 1353 & 1469 & 1100\\
\cline{2-7}
\multicolumn{1}{ |c| }{} & \multicolumn{1}{ |c| }{\multirow{4}{*}{Model 4}} & \multicolumn{1}{ |c| }{RMSE} & 6212.31 & 12706.26 & {32646.63} & 52751.72  \\
\multicolumn{1}{ |c| }{} & \multicolumn{1}{ |c| }{} & \multicolumn{1}{ |c| }{MAPE} & {41.08} & 50.42 & {37.80} & {19.56}  \\ 
\multicolumn{1}{ |c| }{} & \multicolumn{1}{ |c| }{} & \multicolumn{1}{ |c| }{MAE} & 4775.04 & 10641.48 & {26849.83} & {42106.65} \\ 
\multicolumn{1}{ |c| }{} & \multicolumn{1}{ |c| }{} & \multicolumn{1}{ |c| }{Time(s)} & 2154 & 1772 & 2100 & 2254\\
\cline{2-7}
\multicolumn{1}{ |c| }{} & \multicolumn{1}{ |c| }{\multirow{4}{*}{Model 5}} & \multicolumn{1}{ |c| }{RMSE} & 5403.45 & 14679.94 & {36773.00} & 52649.04  \\
\multicolumn{1}{ |c| }{} & \multicolumn{1}{ |c| }{} & \multicolumn{1}{ |c| }{MAPE} & {35.38} & 55.43 & {45.63} & 19.76  \\ 
\multicolumn{1}{ |c| }{} & \multicolumn{1}{ |c| }{} & \multicolumn{1}{ |c| }{MAE} & 4361.95 & 12294.69 & {30163.46} & 42906.34 \\ 
\multicolumn{1}{ |c| }{} & \multicolumn{1}{ |c| }{} & \multicolumn{1}{ |c| }{Time(s)} & 1129 & 945 & 1001 & 1122\\
\cline{2-7}
\multicolumn{1}{ |c| }{} & \multicolumn{1}{ |c| }{\multirow{4}{*}{Model 6}} & \multicolumn{1}{ |c| }{RMSE} & 5621.37 & 13893.23 & \textcolor{red}{32005.10} & 54944.21  \\
\multicolumn{1}{ |c| }{} & \multicolumn{1}{ |c| }{} & \multicolumn{1}{ |c| }{MAPE} & {35.07} & 48.77 & \textcolor{blue}{34.76} & 20.83  \\ 
\multicolumn{1}{ |c| }{} & \multicolumn{1}{ |c| }{} & \multicolumn{1}{ |c| }{MAE} & 4480.75 & 11734.42 & \textcolor{cyan}{26004.11} & {43438.74} \\ 
\multicolumn{1}{ |c| }{} & \multicolumn{1}{ |c| }{} & \multicolumn{1}{ |c| }{Time(s)} & 1122 & 1029 & 983 & 912\\
\cline{2-7}
\multicolumn{1}{ |c| }{} & \multicolumn{1}{ |c| }{\multirow{4}{*}{Model 7}} & \multicolumn{1}{ |c| }{RMSE} & {6515.12} & {15772.53} & 34345.54 & {53317.02}  \\
\multicolumn{1}{ |c| }{} & \multicolumn{1}{ |c| }{} & \multicolumn{1}{ |c| }{MAPE} & {43.30} & {65.44} & 45.34 & 20.82  \\ 
\multicolumn{1}{ |c| }{} & \multicolumn{1}{ |c| }{} & \multicolumn{1}{ |c| }{MAE} & 5208.36 & 13215.27 & 27508.65 & 43108.25 \\ 
\multicolumn{1}{ |c| }{} & \multicolumn{1}{ |c| }{} & \multicolumn{1}{ |c| }{Time(s)} & 987& 998 & 1112 & 1010\\
\cline{1-7}

\multicolumn{1}{c|}{\multirow{28}{*}{\textbf{Cluster 2}}} & \multicolumn{1}{ |c| } {Architecture} & \multicolumn{1}{ |c| }{Performance Measure} &  150 & 200 & 300 & 400 \\ \cline{1-7}

\multicolumn{1}{ |c| }{} & \multicolumn{1}{ |c| }{\multirow{4}{*}{Model 1}} & \multicolumn{1}{ |c| }{RMSE} & {12059.14} & 29017.86 & 47730.52 & 62920.93  \\
\multicolumn{1}{ |c| }{} & \multicolumn{1}{ |c| }{} & \multicolumn{1}{ |c| }{MAPE} & \textcolor{blue}{12.81} & \textcolor{blue}{22.03} & 27.12 & 13.93 \\ 
\multicolumn{1}{ |c| }{} & \multicolumn{1}{ |c| }{} & \multicolumn{1}{ |c| }{MAE} & {8583.00} & {22420.10} & 39235.51 & 51245.27 \\ 
\multicolumn{1}{ |c| }{} & \multicolumn{1}{ |c| }{} & \multicolumn{1}{ |c| }{Time(s)} & 730 & 529 & 625 & 565\\
\cline{2-7}
\multicolumn{1}{ |c| }{} & \multicolumn{1}{ |c| }{\multirow{4}{*}{Model 2}} & \multicolumn{1}{ |c| }{RMSE} & 12281.21 & 28399.50 & 52188.18 & 57151.25  \\
\multicolumn{1}{ |c| }{} & \multicolumn{1}{ |c| }{} & \multicolumn{1}{ |c| }{MAPE} & {15.99} & 22.09 & 19.81 & 11.44  \\ 
\multicolumn{1}{ |c| }{} & \multicolumn{1}{ |c| }{} & \multicolumn{1}{ |c| }{MAE} & 8950.43 & 22083.35 & 42345.36 & 45136.57 \\ 
\multicolumn{1}{ |c| }{} & \multicolumn{1}{ |c| }{} & \multicolumn{1}{ |c| }{Time(s)} & 1048 & 1137 & 1300 & 1323\\
\cline{2-7}
\multicolumn{1}{ |c| }{} & \multicolumn{1}{ |c| }{\multirow{4}{*}{Model 3}} & \multicolumn{1}{ |c| }{RMSE} & \textcolor{red}{11296.55} & \textcolor{red}{26270.88} & {48964.51} & {56242.17}  \\
\multicolumn{1}{ |c| }{} & \multicolumn{1}{ |c| }{} & \multicolumn{1}{ |c| }{MAPE} & {13.36} & {26.19} & {16.36} & {11.75}  \\ 
\multicolumn{1}{ |c| }{} & \multicolumn{1}{ |c| }{} & \multicolumn{1}{ |c| }{MAE} & \textcolor{cyan}{7776.63} & \textcolor{cyan}{19426.56} & {39846.93} & {45253.69} \\ 
\multicolumn{1}{ |c| }{} & \multicolumn{1}{ |c| }{} & \multicolumn{1}{ |c| }{Time(s)} & 1100 & 913 & 848 & 874\\
\cline{2-7}
\multicolumn{1}{ |c| }{} & \multicolumn{1}{ |c| }{\multirow{4}{*}{Model 4}} & \multicolumn{1}{ |c| }{RMSE} & {14693.48} & {31948.36} & \textcolor{red}{46154.93} & 57638.41  \\
\multicolumn{1}{ |c| }{} & \multicolumn{1}{ |c| }{} & \multicolumn{1}{ |c| }{MAPE} & {20.96} & 23.20 & 16.05 & \textcolor{blue}{11.26}  \\ 
\multicolumn{1}{ |c| }{} & \multicolumn{1}{ |c| }{} & \multicolumn{1}{ |c| }{MAE} & {11306.93} & {24519.02} & \textcolor{cyan}{37135.12} & 44802.18 \\ 
\multicolumn{1}{ |c| }{} & \multicolumn{1}{ |c| }{} & \multicolumn{1}{ |c| }{Time(s)} & 1283 & 1525 & 1469 & 1198\\
\cline{2-7}
\multicolumn{1}{ |c| }{} & \multicolumn{1}{ |c| }{\multirow{4}{*}{Model 5}} & \multicolumn{1}{ |c| }{RMSE} & 12000.75 & 33271.13 & 52588.69 & \textcolor{red}{56051.11}  \\
\multicolumn{1}{ |c| }{} & \multicolumn{1}{ |c| }{} & \multicolumn{1}{ |c| }{MAPE} & {14.56} & {25.98} & 18.75 & 11.49  \\ 
\multicolumn{1}{ |c| }{} & \multicolumn{1}{ |c| }{} & \multicolumn{1}{ |c| }{MAE} & 8835.99 & 26458.75 & 42617.07 & \textcolor{cyan}{43542.71} \\ 
\multicolumn{1}{ |c| }{} & \multicolumn{1}{ |c| }{} & \multicolumn{1}{ |c| }{Time(s)} & 664 & 661 & 869 & 867\\
\cline{2-7}
\multicolumn{1}{ |c| }{} & \multicolumn{1}{ |c| }{\multirow{4}{*}{Model 6}} & \multicolumn{1}{ |c| }{RMSE} & 13363.42 & {31299.26} & {47829.81} & 57017.62  \\
\multicolumn{1}{ |c| }{} & \multicolumn{1}{ |c| }{} & \multicolumn{1}{ |c| }{MAPE} & {19.77} & 22.93 & \textcolor{blue}{14.89} & 12.19  \\ 
\multicolumn{1}{ |c| }{} & \multicolumn{1}{ |c| }{} & \multicolumn{1}{ |c| }{MAE} & 10311.96 & 23920.97 & {38111.54} & 45107.26 \\ 
\multicolumn{1}{ |c| }{} & \multicolumn{1}{ |c| }{} & \multicolumn{1}{ |c| }{Time(s)} & 701 & 779 & 923 & 951\\
\cline{2-7}
\multicolumn{1}{ |c| }{} & \multicolumn{1}{ |c| }{\multirow{4}{*}{Model 7}} & \multicolumn{1}{ |c| }{RMSE} & 14424.54 & 30437.25 & 52917.75 & 59926.56 \\
\multicolumn{1}{ |c| }{} & \multicolumn{1}{ |c| }{} & \multicolumn{1}{ |c| }{MAPE} & {17.57} & 22.48 & 17.76 & 11.96 \\ 
\multicolumn{1}{ |c| }{} & \multicolumn{1}{ |c| }{} & \multicolumn{1}{ |c| }{MAE} & 10690.42 & 23874.18 & 43415.56 & 47402.78 \\ 
\multicolumn{1}{ |c| }{} & \multicolumn{1}{ |c| }{} & \multicolumn{1}{ |c| }{Time(s)} & 923 & 1187 & 1015 & 986\\
\cline{1-7}
\end{tabular}
}

\end{table}

According to the Table VI, the total error measures for K=150 for method A  is as follows:
\begin{equation}
    \begin{split}
         RMSE_{Tot}  = \dfrac{142}{400} \times 3975.74 + \dfrac{258}{400}\\ \times 11872.12 = 9068.91
         \end{split}
    \end{equation}
    \begin{equation}
    \begin{split}
         MAPE_{Tot}  = \dfrac{142}{400} \times 32.34 + \dfrac{258}{400} \\ \times 26.55 = 28.61
         \end{split}
    \end{equation}
    \begin{equation}
    \begin{split}
         MAE_{Tot}  = \dfrac{142}{400} \times 3204 + \dfrac{258}{400} \\
         \times 8838.86 = 6838.48
         \end{split}
    \end{equation}
The same calculation can be done according to the Table VII for method B and K=150:
\begin{equation}
    \begin{split}
         RMSE_{Tot}  = \dfrac{230}{400} \times 4906.85 + \dfrac{170}{400}\\ \times 11296.55 = 7622.47
         \end{split}
    \end{equation}
    \begin{equation}
    \begin{split}
         MAPE_{Tot}  = \dfrac{230}{400} \times 31.89 + \dfrac{170}{400} \\ \times 12.81= 23.78
         \end{split}
    \end{equation}
    \begin{equation}
    \begin{split}
         MAE_{Tot}  = \dfrac{230}{400} \times 3591.33 + \dfrac{170}{400} \\
         \times 7776.63 = 5370,08
         \end{split}
    \end{equation}
    
Comparing (7) up to (15) reveal that DTW and method B clustering methods provide the best improvement to prediction Errors. Furthermore, according to Table IV, feature-based clustering methods (method A, method B) significantly outperform the DTW method in terms of speed. Therefore if speed and efficiency are the primary concern, method A and method B can be combined with the neural networks to improve efficiency and error. If improving error is the main objective, then method B and DTW are a clear choice.

Now consider the goal is to compare different clustering algorithms based on the improvement they provide to forecasting errors for all the architectures. This means we are not choosing the best model like before, but we consider all the models for each clustering method. For example, to find the RMSE error for model 1 for DTW clustering, we find the total RMSE when model 1 is used for all the clusters. The results of this analysis for DTW, method A, and method B are summarized in Tables VIII, IX, and X, respectively. 
\begin{table}[tbh]
\caption{Forecasting Results For DTW Clustering For All the Models}
\resizebox{0.5\textwidth}{!}{
\begin{tabular}{ |c|c|c|c|c|c| } 
\hline
Architecture & Performance Measure &  150 & 200 & 300 & 400 \\
\hline
\multirow{3}{*}{Model 1} & RMSE & 7615.73 & 17686.91 & 38501.34 & 56517.08 \\
 & MAPE & {23.41} & 30.45 & 28.50 & 16.89 \\ 
& MAE & 5946.97 & 13581.36 & 31518.53 & 45243.00 \\ 
\hline
\multirow{3}{*}{Model 2} & RMSE & 8158.80 & 17733.35 & 39832.98 & 57018.20 \\
 & MAPE & 21.96 & 31.22 & 27.60 & 16.27 \\ 
& MAE & 6578.64 & {13736.02} & 31260.66 & 45133.94 \\ 
\hline
\multirow{3}{*}{Model 3} & RMSE & {8255.53} & 17507.13 & {41052.15} & {58307.62} \\
 & MAPE & 23.82 & {30.20} & {27.57} & {16.77} \\ 
& MAE & {6260.05} & 14636.79 & 32245.35 & {47219.31} \\ 
\hline
\multirow{3}{*}{Model 4} & RMSE & 8763.41 & {19744.27} & 37456.21 & 62427.74 \\
 & MAPE & 23.26 & 31.60 & 30.44 & 18.98 \\ 
& MAE & 7129.48 & 14797.23 & 31429.43 & 49847.45 \\ 
\hline
\multirow{3}{*}{Model 5} & RMSE & 8373.34 & 19949.35 & 40925.33 & 61215.21 \\
 & MAPE & 23.77 & 34.73 & 29.41 & 18.27 \\ 
& MAE & 6634.74 & 15694.30 & {32983.98} & 49329.41 \\ 
\hline
\multirow{3}{*}{Model 6} & RMSE & 7785.81 & 17495.57 & 39707.54 & 63175.11 \\
 & MAPE & 23.57 & 30.47 & 28.24 & 18.26 \\ 
& MAE & 6201.85 & 12337.61 & 32363.83 & 45967.80 \\ 
\hline
\multirow{3}{*}{Model 7} & RMSE & 11808.89 & 18664.58 & 39572.46 & 59069.45 \\
 & MAPE & 30.02 & 34.88 & 29.44 & 17.22 \\ 
& MAE & 9216.95 & 14718.56 & 32225.28 & 47848.42 \\ 
\hline
\end{tabular}
}

\end{table}

\begin{table}[tbh]
\caption{Forecasting Results For Method A Clustering For All Models}
\resizebox{0.5\textwidth}{!}{
\begin{tabular}{ |c|c|c|c|c|c| } 
\hline
Architecture & Performance Measure &  150 & 200 & 300 & 400 \\
\hline
\multirow{3}{*}{Model 1} & RMSE & 9091.27 & 22106.80 & 45810.48 & 58472.99 \\
 & MAPE & {30.95} & 32.96 & 30.10 & 16.73 \\ 
& MAE & 6838.48 & 17257.40 & 38437.05 & 41366.98 \\ 
\hline
\multirow{3}{*}{Model 2} & RMSE & 10813.55 & 22583.19 & 44151.13 & 55400.72 \\
 & MAPE & 33.23 & 36.81 & 27.99 & 15.55 \\ 
& MAE & 8679.21 & {18376.66} & 38882.59 & 44018.64 \\ 
\hline
\multirow{3}{*}{Model 3} & RMSE & {10473.49} & 22270.56 & {46514.71} & {52870.65} \\
 & MAPE & 34.66 & {35.56} & {31.10} & {14.96} \\ 
& MAE & {8673.15} & 17798.95 & 38882.59 & {42851.77} \\ 
\hline
\multirow{3}{*}{Model 4} & RMSE & 11842.51 & {24389.59} & 47628.19 & 55854.06 \\
 & MAPE & 39.73 & 41.19 & 30.16 & 14.77 \\ 
& MAE & 9810.48 & 19744.74 & 39298.22 & 43544.99 \\ 
\hline
\multirow{3}{*}{Model 5} & RMSE & 10108.43 & 22233.00 & 47295.88 & 58430.44 \\
 & MAPE & 34.51 & 32.64 & 28.56 & 16.99 \\ 
& MAE & 8211.27 & 17594.76 & {37570.27} & 47555.34 \\ 
\hline
\multirow{3}{*}{Model 6} & RMSE & 9236.03 & 22085.80 & 44145.01 & 56069.45 \\
 & MAPE & 28.60 & 40.66 & 29.43 & 15.80 \\ 
& MAE & 7330.97 & 18176.19 & 35406.69 & 44342.42 \\ 
\hline
\multirow{3}{*}{Model 7} & RMSE & 9497.82 & 27619.06 & 49525.77 & 58274.16 \\
 & MAPE & 46.49 &  48.50 & 34.59 & 17.04 \\ 
& MAE & 7409.28 & 23853.25 & 44812.72 & 48237.39 \\ 
\hline
\end{tabular}
}

\end{table}

\begin{table}[tbh]
\caption{Forecasting Results For Method B Clustering For All Models}
\resizebox{0.5\textwidth}{!}{
\begin{tabular}{ |c|c|c|c|c|c| } 
\hline
Architecture & Performance Measure &  150 & 200 & 300 & 400 \\
\hline
\multirow{3}{*}{Model 1} & RMSE & 8725.95 & 20729.13 & 43542.89 & 58448.04 \\
 & MAPE & {25.28} & 34.24 & 39.22 & 17.21 \\ 
& MAE & 6489.63 & 16288.55 & 35578.36 & 46585.77 \\ 
\hline
\multirow{3}{*}{Model 2} & RMSE & 8548.82 & 19254.18 & 41610.98 & 54579.99 \\
 & MAPE & 25.25 & 34.52 & 32.57 &  16.29 \\ 
& MAE & 6489.63 & {15560.71} & 33975.50 & 43712.66 \\ 
\hline
\multirow{3}{*}{Model 3} & RMSE & {7622.47} & 18147.92 & {39783.22} & {52437.68} \\
 & MAPE & 24.01 & {35.53} & {28.77} & {15.03} \\ 
& MAE & {5370.08} & 14110.13 & 32501.80 & {42247.53} \\ 
\hline
\multirow{3}{*}{Model 4} & RMSE & 9816.80 & {20884.15} & 38387.65 & 54828.56 \\
 & MAPE & 32.52 & 38.85 &  28.55 & 16.03 \\ 
& MAE & 7551.09 & 16539.43 & 31221.07 & 43252.25 \\ 
\hline
\multirow{3}{*}{Model 5} & RMSE & 8207.30 & 22581.19 & 43494.66 & 54094.91 \\
 & MAPE & 26.53 & 42.91 & 34.20 & 16.24 \\ 
& MAE & 6263.41 & 18314.41 & {35456.24} & 43176.79 \\ 
\hline
\multirow{3}{*}{Model 6} & RMSE & 8911.74 & 21290.79 & 38730.60 & 55825.40 \\
 & MAPE & 28.56 & 37.78 & 26.31 & 17.15 \\ 
& MAE & 6959.01 & 16913.70 & 31149.76 & 44147.86 \\ 
\hline
\multirow{3}{*}{Model 7} & RMSE & 9876.62 & 22005.03 & 40076.35 & 56126.07 \\
 & MAPE & 32.36 &  47.18 & 32.39 & 17.05 \\ 
& MAE & 7538.23 & 17745.30 & 34269.08 & 44933.42 \\ 
\hline
\end{tabular}
}

\end{table}

Analyzing Tables VIII, IX, and X reveal that DTW clustering provides better overall improvement to MAPE, RMSE and MAE scores when K=150,200, 300, while for K=400 both methods A and method B outperform DTW. However, comparing Tables III, VIII, IX, and X reveal that regardless of the clustering algorithms, forecasting accuracies can significantly be improved when data clustering is done. For example, consider K=300, with no clustering the best MAPE score is 38.52 for model 3; however, with DTW clustering model 3 achieves MAPE score of 27.57, methods A and B achieve MAPE scores of 31.10 and 28.88 respectively. 
\subsection{Influence of adding more Time Series Measurements}
Thus far, we have used the gas measurement along with static data to predict future gas values. However, an important question to ask is, can prediction errors be improved by including more time series measurement?  

To answer this question, we will consider all possible combinations of multivariate time series column and compare their prediction errors to Table III. To show the concept and also save simulation time, the idea is tested only against model 1 and model 3.
\begin{table}
\caption{Influence of Adding more Measurements on Future Gas Value (Third Column) }
\large
\resizebox{0.5\textwidth}{!}{
\begin{tabular}{ |cccc|c|c|c|c|c|c| } 
\cline{2-7}
\multicolumn{1}{c|}{\multirow{12}{*}{\textbf{Model 1}}} & \multicolumn{1}{ |c| } {Measurements Column} & \multicolumn{1}{ |c| }{Performance Measure} &  150 & 200 & 300 & 400 \\ \cline{1-7}
\multicolumn{1}{ |c| }{} & \multicolumn{1}{ |c| }{\multirow{3}{*}{First and Second Columns}} & \multicolumn{1}{ |c| }{RMSE} & 39450.01 & 56527.14 & 84660.19 & 86700.9 \\ 
\multicolumn{1}{ |c| }{} & \multicolumn{1}{ |c| }{} & \multicolumn{1}{ |c| }{MAPE} & {150.38} & 91.89 & 62.33 & 25.86 \\ 
\multicolumn{1}{ |c| }{} & \multicolumn{1}{ |c| }{} & \multicolumn{1}{ |c| }{MAE} & 29850.68 & 44804.45 & 69698.23 & 70742.31 \\ 
\cline{2-7}
\multicolumn{1}{ |c| }{} & \multicolumn{1}{ |c| }{\multirow{3}{*}{First and Third Columns}} & \multicolumn{1}{ |c| }{RMSE} & 10218.19 & 27168.96 & 58130.94 & 72662.22  \\
\multicolumn{1}{ |c| }{} & \multicolumn{1}{ |c| }{} & \multicolumn{1}{ |c| }{MAPE} & {34.36} & 47.14 & 45.89 & 20.66  \\ 
\multicolumn{1}{ |c| }{} & \multicolumn{1}{ |c| }{} & \multicolumn{1}{ |c| }{MAE} & 7122.69 & 19889.56 & 45096.30 & 55307.80 \\ 
\cline{2-7}
\multicolumn{1}{ |c| }{} & \multicolumn{1}{ |c| }{\multirow{3}{*}{Second and Third Columns}} & \multicolumn{1}{ |c| }{RMSE} & 10039.66 & {26831.03} & 60065.97 & {69787.29}  \\
\multicolumn{1}{ |c| }{} & \multicolumn{1}{ |c| }{} & \multicolumn{1}{ |c| }{MAPE} & {34.53} & 43.16 & 41.94 & 21.14  \\ 
\multicolumn{1}{ |c| }{} & \multicolumn{1}{ |c| }{} & \multicolumn{1}{ |c| }{MAE} & 7311.21 & 18740.35 & 46414.14 & 55175.58 \\ 
\cline{2-7}
\multicolumn{1}{ |c| }{} & \multicolumn{1}{ |c| }{\multirow{3}{*}{First, Second and Third}} & \multicolumn{1}{ |c| }{RMSE} & 10319.54 & 26638.20 & 55663.48 & 72562.50  \\
\multicolumn{1}{ |c| }{} & \multicolumn{1}{ |c| }{} & \multicolumn{1}{ |c| }{MAPE} & {36.33} & 42.75 & 45.02 & 20.97  \\ 
\multicolumn{1}{ |c| }{} & \multicolumn{1}{ |c| }{} & \multicolumn{1}{ |c| }{MAE} & 7257.43 & {19944.56} & 49724.80 & 54406.62 \\ 
\cline{1-7}

\multicolumn{1}{c|}{\multirow{12}{*}{\textbf{Model 3}}} & \multicolumn{1}{ |c| } {Measurements Column} & \multicolumn{1}{ |c| }{Performance Measure} &  150 & 200 & 300 & 400 \\ \cline{1-7}
\multicolumn{1}{ |c| }{} & \multicolumn{1}{ |c| }{\multirow{3}{*}{First and Second Columns}} & \multicolumn{1}{ |c| }{RMSE} & 42960.62 & 52977.89 & 94306.01 & 88665.62 \\ 
\multicolumn{1}{ |c| }{} & \multicolumn{1}{ |c| }{} & \multicolumn{1}{ |c| }{MAPE} & {98.84} & 106.53 & 80.84 & 25.47 \\ 
\multicolumn{1}{ |c| }{} & \multicolumn{1}{ |c| }{} & \multicolumn{1}{ |c| }{MAE} & 30140.50 & 41278.36 & 78561.96 & 72679.35 \\ 
\cline{2-7}
\multicolumn{1}{ |c| }{} & \multicolumn{1}{ |c| }{\multirow{3}{*}{First and Third Columns}} & \multicolumn{1}{ |c| }{RMSE} & 10973.27 & 28103.06 & 53204.26 & 71642.94  \\
\multicolumn{1}{ |c| }{} & \multicolumn{1}{ |c| }{} & \multicolumn{1}{ |c| }{MAPE} & {34.33} & 48.45 & 39.75 & 20.41  \\ 
\multicolumn{1}{ |c| }{} & \multicolumn{1}{ |c| }{} & \multicolumn{1}{ |c| }{MAE} & 7886.56 & 20174.22 & 41335.21 & 61775.57 \\ 
\cline{2-7}
\multicolumn{1}{ |c| }{} & \multicolumn{1}{ |c| }{\multirow{3}{*}{Second and Third Columns}} & \multicolumn{1}{ |c| }{RMSE} & 10417.47 & {26390.57} & 55478.94 & 71627.27  \\
\multicolumn{1}{ |c| }{} & \multicolumn{1}{ |c| }{} & \multicolumn{1}{ |c| }{MAPE} & {34.56} & 47.36 & 39.31 & 21.55  \\ 
\multicolumn{1}{ |c| }{} & \multicolumn{1}{ |c| }{} & \multicolumn{1}{ |c| }{MAE} & 7036.22 & 20164.46 & 45503.44 & 55921.33 \\ 
\cline{2-7}
\multicolumn{1}{ |c| }{} & \multicolumn{1}{ |c| }{\multirow{3}{*}{First, Second and Third}} & \multicolumn{1}{ |c| }{RMSE} & 10287.64 & 26195.90 & 63499.57 & 71540.70  \\
\multicolumn{1}{ |c| }{} & \multicolumn{1}{ |c| }{} & \multicolumn{1}{ |c| }{MAPE} & {36.33} & 45.75 & 38.02 & 18.97  \\ 
\multicolumn{1}{ |c| }{} & \multicolumn{1}{ |c| }{} & \multicolumn{1}{ |c| }{MAE} & 7257.43 & {19944.56} & 49724.80 & 52406.62 \\ 
\cline{1-7}
\end{tabular}
}
\end{table}

Comparing Table XI to Table III reveals that adding more time series measurements is not, in general, improving the RMSE, MAPE, and MAE. Such high errors in Table XI corresponding to the first entries of model 1 and model 3 suggest that the first and second columns alone are not good predictors of the third column. 

\section{Conclusion}
In this paper, we provided a comprehensive analysis of time series forecasting using neural networks. We introduced several architectures to combine static and dynamic measurements to forecast time series using neural networks. Our results reveal that: First, not one architecture is right for predicting all future values of time series, but multiple architectures should be tested, and the best architecture should be determined based on model selection criterion such as cross-validation. Second, adding static data to forecast does not necessarily improve the forecasting errors.  

We discussed two primary clustering methods, namely, distance-based and feature-based clustering. We showed that DTW outperforms the feature-based clustering in most architectures; however, feature-based clustering methods outperform the DTW in terms of speed and time complexity and still improve the forecasting accuracies significantly when they are compared to no clustering case. 

Finally, we presented several methods, such as anomaly detection and data imputation, to prepare time series data for forecasting and the clustering.


%





\ifCLASSOPTIONcaptionsoff
  \newpage
\fi



%
\bibliographystyle{IEEEtran}

\bibliography{Bib_tran1}


%








\end{document}